\documentclass{article}

\PassOptionsToPackage{numbers, sort}{natbib}

\usepackage[final]{neurips_2025}

\usepackage[utf8]{inputenc} 
\usepackage[T1]{fontenc}    
\usepackage{hyperref}       
\usepackage{url}            
\usepackage{booktabs}       
\usepackage{amsfonts}       
\usepackage{nicefrac}       
\usepackage{microtype}      
\usepackage{xcolor}         

\usepackage{graphicx}
\graphicspath{ {./figs/} }
\usepackage{placeins}

\usepackage{amsmath,amssymb}
\DeclareMathOperator{\E}{\mathbb{E}}
\DeclareMathOperator{\Var}{\mathrm{Var}}
\DeclareMathOperator{\Cov}{\mathrm{Cov}}

\usepackage{multirow}

\usepackage{amsthm}
\usepackage{mathtools}
\usepackage{relsize}
\usepackage{xfrac}
\usepackage{caption}
\usepackage{soul}
\sethlcolor{green}

\title{Complexity Scaling Laws for Neural Models\\using Combinatorial Optimization}

\author{
    Lowell Weissman\thanks{Bradley Department of Electrical and Computer Engineering. Correspondence to: <lowell6@vt.edu>}
    \\
    Virginia Tech
    \And
    Michael Krumdick\\
    Kensho Technologies
    \And
    A. Lynn Abbott\footnotemark[2] \\
    Virginia Tech
}

\begin{document}

\maketitle

\setcounter{footnote}{0}
\begin{abstract}
    Recent work on neural scaling laws demonstrates that model performance scales predictably with compute budget, model size, and dataset size. In this work, we develop scaling laws based on \textit{problem complexity}. We analyze two fundamental complexity measures: solution space size and representation space size. Using the Traveling Salesman Problem (TSP) as a case study, we show that combinatorial optimization promotes smooth cost trends, and therefore meaningful scaling laws can be obtained even in the absence of an interpretable loss. We then show that suboptimality grows predictably for fixed-size models when scaling the number of TSP nodes or spatial dimensions, independent of whether the model was trained with reinforcement learning or supervised fine-tuning on a static dataset. We conclude with an analogy to problem complexity scaling in local search, showing that a much simpler gradient descent of the cost landscape produces similar trends.\footnote{Project code, along with our new dataset of 128 million optimal TSP solutions, is provided at:\\ \url{https://github.com/lowellw6/complexity-scaling-laws}}
\end{abstract}

\section{Introduction}
\label{intro}
Neural network performance usually improves with more parameters, more training, and more data. While small increases in budget improve performance unpredictably, model performance becomes highly predictable at scale, often with surprisingly simple trends \cite{hestness2017deep, rosenfeld2019constructive}. Such trends have been called neural scaling laws \cite{kaplan2020scalinglawsneurallanguage, henighan2020scaling}, where performance smoothly improves when scaling model size, dataset size, or compute budget. This principle is summarized elegantly \cite{kaplan2020scalinglawsneurallanguage} by analogy to the ideal gas law, which describes the macroscopic behavior of a gas independent of its microscopic dynamics.

Neural scaling laws emerge over a wide variety of problems and problem scales \cite{henighan2020scaling, alabdulmohsin2022revisiting, zhai2022scaling, gao2023scaling, neumann2022scaling, lee2022multi}. This observation implies the existence of a general underlying order, one that persists across the diverse natures of various tasks. Neural scaling laws capture slices of this behavior, fixing variables of the problem so that their influence is absorbed within the constants of a specific trend.

In this paper, we hypothesize that model performance is similarly predictable when scaling fundamental measures of problem complexity. We directly isolate this relationship and scale problem complexity for parameter-constrained, fixed-size models while approximating the regime in which compute and data are unconstrained. We take an initial step toward predicting the limit of performance as a function of the deep learning algorithm, the capacity of the model, and the properties of the task.

If we study measures of complexity that are task-specific, it will be difficult to draw meaningful conclusions that are general in nature, even if smooth scaling laws emerge. For example, Jones~\cite{jones2021scaling} extends compute scaling laws to also be a function of Hex board size. Measures like game-tree complexity and state-space complexity are well-studied for Hex~\cite{van2002games} but have complicated relationships with board size~\cite{van2002computer} and are primarily applicable to sequential games. We approach this issue by distilling two fundamental measures of problem complexity inherent in any deep learning task: size of the solution space and size of the representation space. Isolating each measure to a single variable is often not feasible in prevalent domains such as generative modeling and vision. We leverage combinatorial optimization toward this goal, using the Traveling Salesman Problem (TSP) to decouple the solution space and representation space, scaling TSP nodes or spatial dimensions independently. We then examine the limit of model performance when confronting the combinatorial growth of the solution space along with the curse of dimensionality. 

\begin{figure}[t]
    \centering
    \includegraphics[clip, trim=0.1in 0.1in 0.1in 0.09in, width=5.5in]{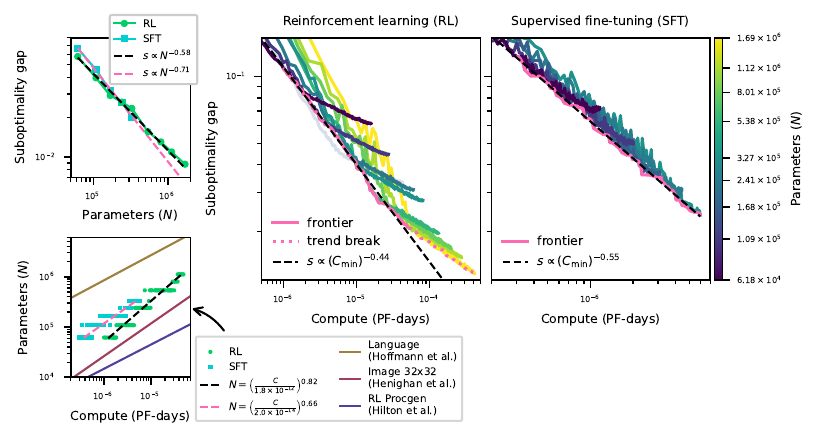}
    \caption{Suboptimality gap, defined as the difference between mean model performance and mean optimal performance, smoothly power decays with respect to model size and compute for both reinforcement learning (RL) and supervised fine-tuning (SFT) in TSP. Fits suggest that SFT is more compute-efficient than RL, and possibly more parameter-efficient as we scale to larger models, with faster decay toward optimal performance (larger \(\alpha\)). \textbf{Top left:} Suboptimality w.r.t. parameters (\(N\)) for models evaluated near convergence. \textbf{Right:} Suboptimality w.r.t. compute (\(C\)) evaluated throughout training, where the compute-efficient frontier power decays. Note that the compute axis for SFT has been stretched for easier viewing. \textbf{Bottom left:} Optimal model size follows power growth w.r.t. compute budget. This relationship is strikingly consistent between domains~\cite{hoffmann2022training, henighan2020scaling, hilton2023scaling}.}
    \label{fig:sjoint}
\end{figure}

\textbf{Contributions:} We obtain model size and compute budget scaling laws for TSP that predict model suboptimality, allowing us to make direct comparisons between reinforcement learning (RL) and supervised fine-tuning (SFT) on the same task (Section~\ref{capacity}; Figure \ref{fig:sjoint}). We then obtain smooth problem complexity scaling laws characterized by simple trends for parameter-constrained models (Section~\ref{pcs}). However, we show that these trends must break down at very large TSP node scales, reaffirming~\cite{kaplan2020scalinglawsneurallanguage} that simple scaling behavior can become more nuanced beyond the scales studied. Informed by comparisons to local search, we provide potential explanations for our parameter-constrained complexity scaling laws (Section~\ref{searchreproduce}). Finally, we provide our newly created dataset of 128 million TSP solutions.

\section{Experimental setup}
\label{method}

\subsection{Data design}
\label{datadesign}
\begin{figure}[t]
    \centering
    \includegraphics[clip, trim=0.1in 0.1in 0.1in 0.1in, width=5.5in]{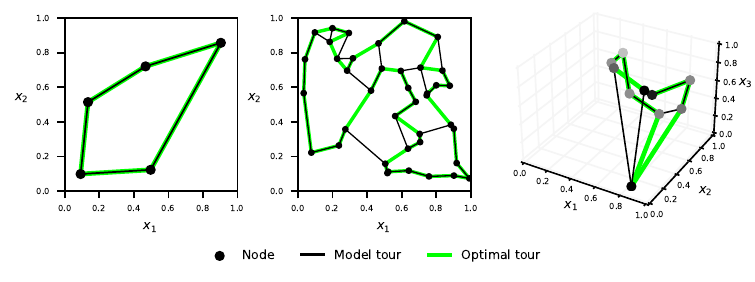}
    \caption{TSP has two convenient ways to adjust problem complexity: node count and spatial dimensionality. \textbf{Left:} 2D TSP instance with 5 nodes and a trivial 12 solutions. The solution tour sampled from a trained RL model is the optimal, minimum-length tour. \textbf{Center:} 2D TSP instance with 40 nodes and roughly \(10^{46}\) solutions. The RL model tour is slightly suboptimal, 0.08 units longer than the optimal tour. \textbf{Right:} 3D TSP instance with 10 nodes, where brightness illustrates increased depth. Adding spatial dimensions does not modify the number of solutions but makes the problem representation more complex. This RL model solution is 0.05 units suboptimal.}
    \label{fig:cover}
\end{figure}

All experiments were performed using the Symmetric Euclidean Traveling Salesman Problem. One problem instance consists of \(n\) \(\mathbb{R}^d\) coordinates uniformly sampled between \([0,1]\) for each of the \mbox{\(d\)}~dimensions (illustrated in Figure \ref{fig:cover}). TSP's \(O(n!)\) combinatorial solution space is defined as all closed loops that visit each node exactly once. Such loops are called \textit{tours}, and the objective is to find the minimum-length tour. In combinatorial optimization language, tour length is referred to as \textit{cost} (or \textit{fitness} when maximizing), and the cost surface w.r.t. the solution space is referred to as the cost \textit{landscape}.

\subsection{Optimal tour generation}
\label{optsolgen}
We leverage optimal solutions for model evaluation and supervised model training. For two-dimensional TSP, we use PyConcorde~\cite{pyconcorde}, a Python wrapper for the Concorde TSP solver~\cite{applegate1998solution, concordeweb}. We generate optimal solutions to 128 million TSP problems, 12.8 million for each node scale we study, and share this dataset for academic use. We were unable to identify a pre-existing dataset of adequate size for supervised learning.

For our evaluations on higher-dimensional TSP, we closely approximate optimality with local search. We discuss our method and how we validated the resulting datasets in Appendix~\ref{apx:proxyval}. Although not exact, performance estimates derived from these datasets are statistically indistinguishable from their Concorde counterparts at two dimensions. We produce 128,000 of these solutions for each 10-node dimension scale we study, and 64,000 solutions for each 20-node dimension scale.

\subsection{Model training}
We train models separately using both reinforcement learning (RL) and supervised fine-tuning~(SFT). Our models output tour sequences autoregressively, producing a policy distribution over each unvisited node via a Pointer Network head~\cite{vinyals2015pointer}, adding an encoding of the sampled visitation to its Transformer decoder memory~\cite{vaswani2017attention}, and repeating until all nodes are visited. Details on our model architecture are provided in Appendix~\ref{apx:arch}, along with a brief comparison to alternative approaches. Hyperparameter choices are detailed separately in Appendix~\ref{apx:hpo}.

\textbf{Reinforcement learning:} We formalize the RL environment as a bandit problem where return is the negative length of a solution tour. We use Proximal Policy Optimization~(PPO)~\cite{schulman2017proximalpolicyoptimizationalgorithms}, training for one million gradient updates with hyperparameter optimization~(HPO) informed learning rate decay.

\textbf{Supervised fine-tuning:} For SFT models, we minimize negative log-likelihood (NLL) loss between model edge selections and optimal edge selections with teacher forcing. We train for one epoch over the optimal solution dataset generated for the corresponding problem scale.\footnote{Multiple epochs of training is a distinct scaling regime where data is also bottlenecked  (Eq. 1.5 in~\cite{kaplan2020scalinglawsneurallanguage}).} One epoch translates to roughly 73,000 gradient updates with a faster learning rate decay than that used for RL experiments.

\subsection{Suboptimality estimation}
\label{suboptestimation}
We measure model performance in mean suboptimal tour length, \(s = \mu_{model} - \mu_{opt}\), where \(\mu_{model}\) is mean model tour length and \(\mu_{opt}\) is mean optimal tour length. We refer to this regret-based metric as the \textit{suboptimality gap}, or simply \textit{suboptimality}. For neural scaling laws, raw tour length trends and suboptimality trends both describe the same relationship since \(\mu_{opt}\) remains constant. For problem complexity scaling \(\mu_{opt}\) varies, so measuring suboptimality extracts the underlying model behavior.

Parameter and node scaling evaluations use the first 1.28 million problems of their corresponding Concorde-solved datasets. Compute scaling evaluations, performed throughout training every 4000 updates, use the first 12,800 problems. Spatial dimension scaling evaluations use their full approximate dataset for each scale. We sample one model tour for each optimal tour.

\subsection{Scaling experiments}
\label{scalingmethod}
\textbf{Neural scaling:} For parameter and compute scaling laws, we study two-dimensional 20-node TSP. We train at 12 model sizes, scaling model width to achieve a geometric progression between roughly 60,000 and 6 million parameters. Previous work suggests that neural scaling is relatively insensitive to Transformer width/depth aspect ratio over several orders of magnitude~\cite{kaplan2020scalinglawsneurallanguage}. We would expect similar results if, for instance, the number of layers were scaled instead, provided models are still trained sufficiently near convergence. However, for some tasks such as neural machine translation, proportionality of encoder and decoder blocks must also be considered~\cite{ghorbani2021scaling}.

\textbf{Problem complexity scaling:} For TSP node and spatial dimension scaling laws, we fix model width thereby fixing model capacity as we scale problem complexity. For node scaling, we train one RL and one SFT model per node scale. We study ten scales, \mbox{\(n\in\{5, 10, 15, ..., 50\}\)}. Symmetric TSP has \mbox{\(\sfrac{1}{2}\,(n-1)!\)} possible tours\footnote{From \mbox{\(n!\)} permutations, starting node insensitivity yields \mbox{\(n\)} equivalent tours per permutation and tour reversal insensitivity yields \mbox{\(2\)} equivalent tours per permutation (dividing by \mbox{\(n\)} and \mbox{\(2\)}, respectively).} for \mbox{\(n\ge3\)}, so these scales range from 12 solutions up to approximately \mbox{\(3 \times 10^{62}\)}. For spatial dimension scaling, we train one RL model per scale and forgo SFT training given the scarcity of optimal data as detailed in Section~\ref{optsolgen}. We study \mbox{\(d\in\{2, 3, ..., 12, 15, 20, 30, 40, 50, 100\}\)} over two node scales \mbox{\(n\in\{10, 20\}\)}, which allows us to better interpret otherwise ambiguous trends.

\begin{figure}
    \centering
    \includegraphics[clip, trim=0.1in 0.1in 0.1in 0.1in, width=5.5in]{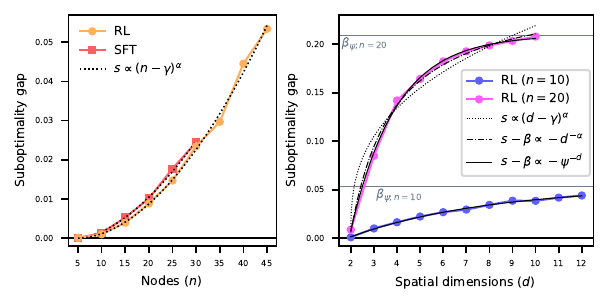}
    \caption{Suboptimality over problem scaling for models near convergence with a fixed number of non-embedding parameters. \textbf{Left:} Suboptimality follows superlinear power growth w.r.t. nodes, though we expect this trend eventually to break down before intersecting the near-linear random performance ceiling (Figure~\ref{fig:bounds}). \textbf{Right:} Suboptimality smoothly increases w.r.t. spatial dimensions, closely following negative exponential decay. Power growth (dashed), power decay (dash-dot), and exponential decay (solid) all predict the 10-node RL experiment (we show the latter). But power growth fails to find a convincing fit for its 20-node counterpart. Even random tour suboptimality is bounded as \(d \to \infty\) (Theorem~\ref{proof:constspan}), so \textit{any} better-than-random monotonic trend must converge. But the power decay asymptote obtained for 10 nodes is larger than that for 20 nodes, which is nonsensical. Exponential decay is most predictive while maintaining sound \(\beta_{\psi}\) asymptote ordering, as shown.}
    \label{fig:complexity}
\end{figure}

\textbf{Fitting scaling laws:} Figures \ref{fig:sjoint} and \ref{fig:complexity} show experiments that converge to trend. These results are the inputs for scaling law regression fits. We discuss experiments that failed to converge in Appendix~\ref{apx:trendbreak}. Critically, all trailing trend-breaking experiments are more suboptimal than predicted. If any were less suboptimal, we could not hypothesize that trend alignment is achievable with further training and further model improvement. Note that RL converges to trend at larger scales than SFT since the former is not constrained by the amount of available optimal data.

\textbf{Compute requirements:} Our 50-node RL experiment is the most computationally expensive. It trained for 24 V100-days and consumed roughly \mbox{\(3 \times 10^{-3}\)} PF-days of compute. Including preliminary experiments and HPO, training required several months. Further details are provided in Appendix~\ref{apx:compute}.

\section{Neural scaling laws for combinatorial optimization}
\label{capacity}
Figure \ref{fig:sjoint} shows that model suboptimality smoothly power decays w.r.t. parameters or compute, for both RL and SFT with the chosen settings. At the scales tested, SFT is more compute efficient than RL (at the expense of node-level supervision), but both exhibit similar parameter efficiency. Note that SFT's parameter-scaling fit may slightly underestimate the true infinite-compute decay exponent. For the larger models shown in SFT's compute-efficient frontier, learning rate decay forces convergence before suboptimality visibly deviates from the frontier. Hence, further training may yield small improvements for these models, increasing SFT's parameter-scaling exponent and further differentiating it from RL's. This suggests SFT will surpass RL in parameter efficiency with larger models. However, with only one seed per model size, we cannot determine statistical significance.

Figure \ref{fig:sjoint} also shows that optimal model size can be expressed as a function of compute budget by extracting each model's contribution to the compute-efficient frontier. This scaling law is very consistent between domains such as generative modeling~\cite{kaplan2020scalinglawsneurallanguage, henighan2020scaling, hoffmann2022training} and reinforcement learning~\cite{hilton2023scaling, neumann2022scaling}. Growth rate exponents are usually around 0.5 to 0.75, and our TSP results closely align with these values. However, the degree of overlap between adjacent model sizes produces non-trivial imprecision, especially for SFT, so extrapolation would be imprecise. We can infer that RL and SFT are compute efficient at similar model sizes for the tested range and TSP problem scale.

A broader implication of Figure \ref{fig:sjoint} is that TSP tour length is a natural performance metric \cite{hilton2023scaling}. For TSP, RL obtains smooth scaling laws directly for the return signal with high precision. For more complex environments with sequential dynamics, return-based RL scaling laws have considerable variance and lose predictive power~\cite{team2023human, lee2022multi, sartor2024neural, tuyls2023scaling}.

Note that the RL compute-efficient frontier ignores the 168,000 parameter model, shown in lower opacity, which improves slightly faster than the trend otherwise predicts. We discuss this outlier in Appendix~\ref{apx:trendbreak}. We include scaling results for loss in Appendix~\ref{apx:loss} and only summarize those findings here. First, loss variance can reveal a pattern of smooth power decay even when the corresponding mean has no trend. Second, PPO's mean critic loss predictably improves despite the absence of a mean actor loss trend and the interdependence between the two learning objectives.

\section{Complexity scaling laws at the infinite-compute limit}
\label{pcs}
We now study problem complexity scaling for deep models with unbottlenecked compute and data, observing patterns in the limit of performance under fixed model capacity. We find that predictable suboptimality trends emerge when scaling either the number of nodes or the number of spatial dimensions, despite the stark contrast in how these measures influence the problem. Further, our node scaling result fosters a useful critique of scaling law breakdown, and our dimension scaling result characterizes the distinct nature of embedding parameters in complexity scaling laws. We again discuss corresponding loss trends in Appendix~\ref{apx:loss}. Theorems introduced are proven in Appendix~\ref{apx:proofs}.

\subsection{Solution space scaling}
\label{solscale}
The left plot in Figure~\ref{fig:complexity} shows suboptimality for RL and SFT models near convergence as we scale the number of TSP nodes. Both experiments exhibit smooth power growth after including a third fitted constant \(\gamma\) as a node-scale offset.\footnote{A node-scale offset is required because zero-node TSP is ill-defined.} We find that growth rates for RL (\(\alpha \approx 1.86\)) and SFT (\(\alpha \approx 1.69\)) are close for the tested model size, although their precise values are not necessarily meaningful. Power growth fits are sensitive to the suboptimalities obtained for the largest node scales, which require the most training to converge. If further training can produce small improvements at these scales, a fit may overestimate the true infinite-compute \(\alpha\). Similar to parameter scaling fits in Figure~\ref{fig:sjoint}, SFT's node-scaling fit is more likely to be pessimistic than RL's given the former's shortened training. However, extrapolating SFT's small \(\alpha\) advantage is less trustworthy here since SFT does not surpass RL by 30 nodes.

\begin{figure}[t]
    \centering
    \includegraphics[clip, trim=0.1in 0.1in 0.1in 0.05in, width=5.5in]{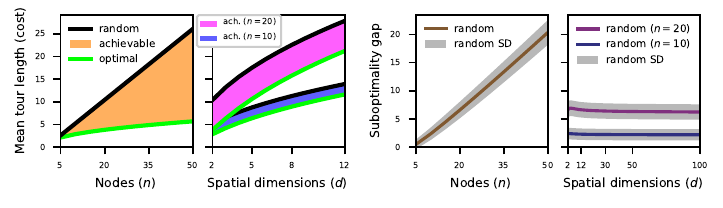}
    \caption{Node and spatial dimension scaling have distinct effects on the achievable performance span, the suboptimality gap of random performance. \textbf{Left:} Mean optimal tour length closely follows sublinear power growth w.r.t. either problem scale. Mean random tour length grows linearly w.r.t. nodes, and sublinearly w.r.t. dimensions at rate similar to optimal tour length growth. Each sublinear trend approaches square root growth in the limit (\(\alpha = 0.5\); proof in Appendix~\ref{apx:proofs}). \textbf{Right:} Suboptimality of random performance w.r.t. nodes is polynomial but approximately linear, being dominated by random tour length growth. Suboptimality of random performance w.r.t. dimensions produces a small, transient increase then decrease, but is provably constant in the limit (Theorem~\ref{proof:constspan}).}
    \label{fig:bounds}
\end{figure}

Even if we assume that \(\alpha\) estimates are imprecise, the growth of suboptimality is clearly superlinear. This observation presents a subtle contradiction. Figure \ref{fig:bounds} shows that the suboptimality of random tour length grows almost linearly with node scale. Indeed, this growth is provably \(O(n)\) as \(n\to\infty\).\footnote{Lemma \ref{proof:expn} shows that expected random tour length grows linearly w.r.t. \(n\), while the Beardwood-Halton-Hammersley Theorem \cite{beardwood1959shortest} proves that optimal tour length is asymptotically proportional to \(\sqrt{n}\) as \(n \to \infty\).} Thus, the obtained scaling laws imply model performance will eventually be worse than random performance, and with enough scale, exceed even the maximum possible tour length (also \(O(n)\)). Instead, we expect this superlinearity to break down before exceeding random performance, and we expect the obtained scaling laws to become pessimistic estimates at large node scales.

Model suboptimality may be better expressed as a broken scaling law \cite{caballero2023broken} given the strong fit quality over initial scaling and the subsequent need for an inflection point. However, this functional form would increase the number of fitted constants, and piecewise scaling is less useful unless we can also predict where trend breaks occur. Conveniently, extrapolating RL's more pessimistic fit, model suboptimality does not intersect random tour suboptimality until roughly 40,000 nodes, a scale with an unfathomably large solution space. Analogous contradictions with huge model sizes have been observed for neural scaling laws (e.g., Figure 15 in \cite{kaplan2020scalinglawsneurallanguage}), reinforcing the caveat that more intricate scaling behavior can appear simple over the scales studied.

\subsection{Representation space scaling}
\label{repscale}
The right plot in Figure~\ref{fig:complexity} shows suboptimality for RL models near convergence as we scale the number of TSP spatial dimensions. Both 10-node and 20-node experiments produce smooth bounded growth described by negative exponential decay toward a fitted asymptote \(\beta\). We found that almost any regression model results in a good fit of the 10-node experiment since its growth is closer to being linear, but the 20-node experiment is more discriminative. Unbounded power growth produces a markedly poor fit given the observed trend is visibly convergent. Power decay regression obtains smaller residuals but appears to converge slightly slower than the observed growth. Further, power decay fits produce a likely contradiction: the bound obtained for the 10-node experiment (\(\beta_{\alpha} \approx 0.31\)) is higher than that obtained for the more complex 20-node counterpart (\(\beta_{\alpha} \approx 0.25\)). This implies that a lower suboptimality is achieved in the limit as \(d\to\infty\) by raising the baseline solution complexity, unless increased node scale fundamentally alters the form of scaling w.r.t. dimensions. In contrast, we find that exponential decay is most predictive\footnote{We also tested decays with the quasi-polynomial form \(d^{-\log_{\psi}d}\) and the subexponential form \(\psi^{{-d}^{\phi}}\), \mbox{\(\phi \in (0,1]\)} but found extreme log base fits for the former and no visual improvement from either form.} while maintaining sound asymptote ordering between 10- and 20-node experiments (the \(\beta_{\psi}\) values shown). Like \(\alpha\) in node scaling fits, \(\beta_\psi\) is sensitive to the suboptimality values obtained at larger scales. Because higher-dimensional runs require more training to converge, \(\beta_\psi\) estimates are likely imprecise, so the exact values shown may not be meaningful.

To explain the asymptotic nature of this scaling law, we again refer to analysis. \(L^p\)-norms such as Euclidean distance exhibit quite unintuitive behavior when scaling to higher dimensions~\cite{aggarwal2001surprising, franccois2007concentration, biau2015high, mirkes2020fractional}. In expectation, both random tour length and optimal tour length diverge but at similar rates (Figure~\ref{fig:bounds}), so random tour suboptimality is roughly constant over the tested domain and is provably constant in the limit as \(d \to \infty\) (Theorem~\ref{proof:constspan}). Several other properties of the cost landscape similarly converge as \(d\to\infty\) (Theorem~\ref{proof:limvard} and Appendix~\ref{apx:landscape}). For local search algorithms, these observations imply that TSP approaches a mostly (if not entirely) stationary problem complexity with one critical exception: the increasing computational complexity of evaluating scalar distances in higher dimensions. (However, the average number of distances that search evaluates converges.)

This increasing computational complexity is confined to the embedding layer for deep models. If we assume that arbitrarily large feature vectors can be embedded without producing a learning bottleneck, then model performance converging to a better-than-random suboptimality becomes anticipated. This is a complementary view of a pattern previously established in neural scaling: the separability of embedding parameters from model capacity~\cite{kaplan2020scalinglawsneurallanguage, henighan2020scaling, hilton2023scaling}. When embeddings are unbottlenecked for a fixed problem, scaling non-embedding parameters studies the bottleneck of model capacity (parameter scaling laws). When scaling the problem, embeddings must also be scaled such that model capacity remains the only bottleneck (parameter-constrained complexity scaling laws).

\section{Interpretable complexity scaling with local search}
\label{searchreproduce}
We now reverse engineer the complexity scaling laws introduced in Section \ref{pcs} using local search, studying comparable trends produced with a white-box algorithm. These similarities do \textit{not} directly imply that model inference and local search share underlying mechanisms. Instead, they provide potential topics to explore toward the development of a formal theory for parameter-constrained deep learning in future work. We evaluate local search using the same datasets discussed in Section \ref{suboptestimation}. For each problem, we initialize search with a random tour and then generate a locally optimum tour with the simple, well-studied 2-opt search move \cite{croes1958method}, performing gradient descent through the cost landscape. The landscape properties discussed at a high level below are detailed in Appendix~\ref{apx:landscape}.

Figure~\ref{fig:jointsearch} shows the suboptimality of 2-opt solutions w.r.t. TSP node and spatial dimension scaling. For the latter, extending to 100 dimensions reveals a clearly convergent trend that aligns with the reasoning introduced in Section \ref{repscale}. If we only observe the lower-dimensional scales studied in Figure~\ref{fig:complexity}, convergence is less obvious, as it is for RL. We find that subexponential decay produces a better fit than pure exponential decay, though at the expense of another fitted constant~\(\phi\). This form may also generalize better for deep models, although we observe no improvement when fitting subexponential decay to the RL experiments in this paper.

Before trends converge, scaling spatial dimensions increases the density of local optima in the cost landscape. In expectation, search arrives at a local optimum faster, but one with increased suboptimality. This trend mirrors the result of parameter-constrained models. 10-node experiments are especially similar, where 2-opt's \(\beta\) asymptote is nearly aligned with RL's at the tested model size. One potential explanation is that fixed model capacity limits a model's ability to avoid poor local optima, matching the bottleneck observed with local search. For example, if models learn a latent representation of a smoothed cost landscape~\cite{gu1994efficient}, more model capacity may permit more advanced smoothing, making low cost optima easier to identify.

When scaling nodes, 2-opt's baseline behavior is less informative. We observe near-linear growth with an inflection point, an unclear relationship that deviates from the power growth observed for RL and SFT. One key distinction between dimension scaling and node scaling is the depth of search required to reach a local optimum. When scaling dimensions before trend convergence, the required search depth slightly decreases due to increased local optima density. In contrast, as \(n\to\infty\), the required search depth diverges to infinity. Attempting to align node scaling behavior, we constrain the maximum search depth (\(M\)). Doing so causes suboptimality to follow smooth superlinear power growth after search saturates at 100\% early stopping (Figure~\ref{fig:jointsearch} top-right and bottom). Although \(M\)-constrained suboptimality is much higher than model suboptimality, and we expect breakdown of superlinearity (before exceeding random tour suboptimality) to occur more rapidly.

Despite these distinctions, the power growth form of \(M\)-constrained suboptimality may still help explain this form for parameter-constrained models. As the cost landscape grows, fixed-depth local search stops at solutions that are further from the attractor local optimum. Is inference similarly forced to approximate local optima under constrained model capacity? If so, what measure of solution proximity is learned? We were able to reproduce superlinear power growth with \(M\)-constrained 2-exchange search, which uses a different definition of solution adjacency (shown in Appendix~\ref{apx:landscape}). Hence, this finding is not overly sensitive to the definition of solution proximity.

\begin{figure}
    \centering
    \includegraphics[clip, trim=0.1in 0.12in 0.1in 0.1in, width=5.5in]{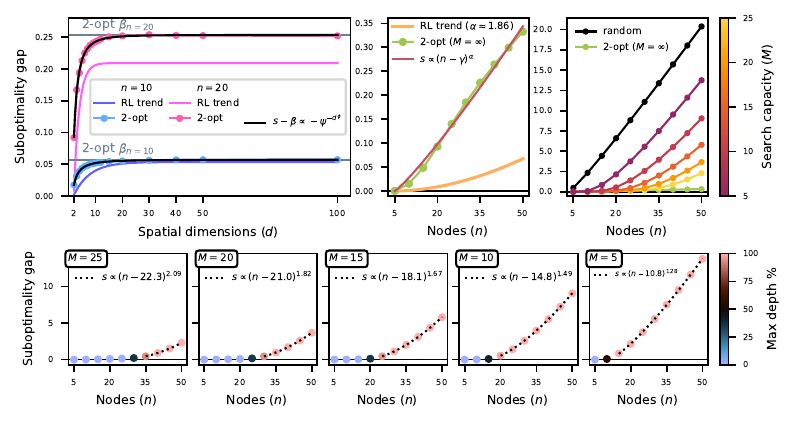}
    \caption{2-opt local search suboptimality over problem complexity scaling. A simpler gradient descent of the cost landscape can produce trends similar to those of parameter-constrained deep models. \textbf{Top~left:} 2-opt suboptimality w.r.t. spatial dimensions closely aligns with RL trends. Pure exponential fit attempts decay slightly too fast, but we obtain close fits with the subexponential generalization shown, where \mbox{\(\phi \in (0,1]\)}. \textbf{Top center:} 2-opt suboptimality w.r.t. number of nodes. With unconstrained search depth, 2-opt produces an unclear trend with an inflection point. \textbf{Top right:} Contraining search depth (\(M\)) produces smooth superlinear growth. \textbf{Bottom:} Power growth emerges after saturating at 100\% early stopping, aligning with the scaling form of parameter-constrained deep models (but these trends are \textit{not} roughly equivalent, because proportionalities are quite different).}
    \label{fig:jointsearch}
\end{figure}

\section{Related work}
\label{relwork}
\paragraph{Neural scaling laws for loss:} Neural scaling laws where cross-entropy loss smoothly improves with model size, dataset size, or compute budget were popularized by Kaplan et al.'s study of large language models~\cite{kaplan2020scalinglawsneurallanguage}, which also first demonstrated the separable nature of embedding parameters. Henighan et al.~\cite{henighan2020scaling} generalize neural scaling laws to other autoregressive generative modeling domains, showing that optimal model size w.r.t. compute budget is remarkably consistent. Sorscher et al.~\cite{sorscher2022beyond} demonstrate that an exponential decay of test error w.r.t. data can be achieved using data pruning (instead of the typical power law scaling~\mbox{\cite{kaplan2020scalinglawsneurallanguage, alabdulmohsin2022revisiting, zhai2022scaling}}). Analogously, Frantar et al.~\mbox{\cite{frantar2024scaling}} show that network sparsity influences scaling behavior. Supervised learning constitutes the majority of recent research on neural scaling laws~\cite{kaplan2020scalinglawsneurallanguage, henighan2020scaling, ghorbani2021scaling, sorscher2022beyond, frantar2024scaling, hernandez2021scaling, alabdulmohsin2022revisiting, zhai2022scaling, hernandez2022scaling, clark2022unified}.

\paragraph{Neural scaling laws for RL:} Reward-based metrics for reinforcement learning models often do not improve smoothly when scaling parameters, compute, or environment interactions~\cite{hilton2023scaling}. Notable exceptions include ground-truth reward in RL from human feedback (RLHF)~\cite{gao2023scaling} and Elo ratings in two-player competitive zero-sum games such as Hex~\cite{jones2021scaling}, Connect Four and Pentago~\cite{neumann2022scaling, neumann2024alphazero}, and to a lesser extent in two-team competitive games like football~\cite{liu2022motor}. Hilton et al.~\cite{hilton2023scaling} address this challenge by introducing intrinsic performance, which maps the compute-efficient frontier to a linear relationship. Caballero et al.~\cite{caballero2023broken} use broken neural scaling laws to predict more sophisticated return trends, like those observed in Procgen tasks~\cite{cobbe2020leveraging}. We circumvent these challenges entirely because TSP natively exhibits smooth, pure power scaling of suboptimality w.r.t. parameters or compute.

\paragraph{Neural scaling law theory:} Several theories have been proposed to explain the relationships between loss and parameters~\cite{sharma2022scaling}, between loss and data~\cite{hutter2021learning}, and for both in conjunction~\cite{bahri2024explaining, michaud2023quantization}. Most recently, Bordelon et al.'s dynamical mean
field theory~\cite{bordelon2024dynamical} recovers phenomena from parameter, data, and compute scaling. The Quantization Hypothesis~\cite{michaud2023quantization, neumann2024alphazero} also explains emergent abilities~\cite{wei2022emergent} by positing that there is an underlying discreteness to all abilities that deep models learn. However, Schaeffer et al.~\cite{schaeffer2023emergent} show that emergent abilities often reflect the chosen performance metric rather than deep learning itself.

\paragraph{Problem complexity scaling laws:} Jones~\cite{jones2021scaling} shows that the Elo score of a fixed-size AlphaZero~\cite{silver2017mastering} model playing Hex can be predicted as a function of both compute budget and board size. The latter primarily grows the solution space similar to TSP node scaling, arguably making that work the closest to our own. However, our key contribution is studying the bottleneck of fixed model capacity, which Jones' method cannot address because it achieves perfect play at each board size. More generally, when considering non-deep-learning algorithms applied to combinatorial optimization, the study of performance w.r.t. problem complexity goes back decades~\cite{smith2012measuring}. Merz et al.~\cite{merz2000fitness} predict the performance of memetic algorithms on the Quadratic Assignment Problem (QAP) using cost landscape properties. (TSP is a special case of QAP.) Ochoa et al.~\cite{ochoa2014local} predict local search performance on NK problems by introducing Local Optima Networks (LON) as a general model of cost landscapes. Tayarani et al.~\cite{tayarani2016analysis} broadly investigate local search trends for TSP, including several node scaling results for cost, some of which informed our methodology.

\section{Discussion}
\label{discussion}
\paragraph{Limitations:} Infinite-compute performance can only be estimated. With modest resources, we could not experiment on larger models and problem scales. This is especially important to consider when extrapolating node scaling trends, since they must eventually break down (possibly well before the contradiction detailed in \mbox{Section~\ref{solscale}}). We are also limited in precision when comparing SFT and RL scaling law fits given that fewer SFT models converge to trend with one epoch of training, and we train only a single seed per scale.\footnote{We provide leave-one-out fit statistics in Appendix~\mbox{\ref{apx:fits}}.} Nevertheless, the main limitation of this work is its focus on a single problem. Euclidean TSP is easier than most NP-hard problems, in part due to its cost metric satisfying the triangle inequality~\mbox{\cite{smith2012measuring}}. Euclidean TSP also admits polynomial-time approximation schemes (PTAS), allowing near-optimal tours to be computed in polynomial time with respect to the number of nodes~\mbox{\cite{arora1998polynomial}}. These properties should be considered when hypothesizing how our findings generalize to other combinatorial optimization problems and beyond. Future work would need to demonstrate compelling generalization before complexity scaling laws can be substantiated as a general principle.

\paragraph{Beyond TSP scaling:} Despite TSP's relative simplicity, there is some basis to be optimistic for parameter-constrained complexity scaling laws in other tasks. The TSP cost landscape was once widely accepted to be approximately globally convex~\cite{boese1994new, angel2001landscape}, but more recent work has found this convexity to be increasingly coarse \cite{hains2011revisiting, ochoa2016multi, ochoa2016deconstructing}. Even so, in expectation, smooth complexity scaling laws emerge. This macroscopic order despite microscopic disorder mirrors the nature of neural scaling laws and supports the hypothesis that complexity scaling laws can be obtained for less structured tasks. A natural next step would be studying parameter-constrained complexity scaling with more elaborate Euclidean combinatorial optimization problems where the solution space and the representation space remain independently scalable. Potential testbeds include other routing problems like the Vehicle Routing Problem~(VRP)~\mbox{\cite{braekers2016vehicle}}, the Orienteering Problem~(OP)~\mbox{\cite{gunawan2016orienteering}}, and the Ring Star Problem~(RSP)~\mbox{\cite{labbe2004ring}}. Given the similarity between the RL node scaling curve and its SFT counterpart (Figure~\mbox{\ref{fig:complexity}} left), we especially encourage future study of algorithm insensitivity for parameter-constrained complexity scaling in other problems.

\paragraph{Real-world domains:} Identifying scalable yet precise measures of problem complexity is very challenging in domains like language modeling, vision, robotics, and so on. Solution complexity and representation complexity are often tightly coupled. For instance, in language modeling, vocabulary size affects both the number of token representations and the space of possible texts to generate. Discriminative NLP tasks like multiple choice question answering maintain separability of vocabulary size and number of classes, but at the expense of diminished relevance to generative modeling. Another challenge arises for real-world tasks where relevant performance metrics do not smoothly improve, as is often the case for sequential decision processes and tasks with discontinuous performance metrics. However, the latter has recently been addressed for language modeling with Token Edit Distance~\cite{schaeffer2023emergent}, an edit-space analogue to~suboptimality.

\paragraph{Complexity scaling law theory:} Without an underlying theory that explains parameter-constrained complexity scaling laws, we can only obtain a limited understanding of the conditions in which they apply and the types of problems they generalize to. Relating findings between domains will require significant speculation and extrapolations may be unreliable. Explaining each form of scaling is the most obvious gap. We are especially interested in whether model size predicts suboptimality growth rates like \(\alpha\) and \(\psi\), or \(\beta\) limits for spatial dimension scaling, and explaining the underlying nature if so. Extending Section~\ref{searchreproduce} to reverse engineer scaling laws with a white-box machine learning model is one possible approach. Identifying and explaining counter-examples is another potentially useful approach. For example, if network sparsity were to influence parameter-constrained complexity scaling laws (analogous to how sparsity influences neural scaling laws~\mbox{\cite{frantar2024scaling}}), a theory explaining this distinction may also address related open questions. Separately, another gap in our current understanding is what determines an embedding bottleneck. Widening the coordinate projection layer forever is probably insufficient to avoid an embedding bottleneck while dimension scaling. Extending the positional encoding sequence forever may be insufficient to avoid one while node scaling.

\paragraph{Practical implications:} Deep methods that excel at solving TSP usually excel at solving similar routing problems like VRP and OP~\mbox{\cite{kool2018attention, kool2022deep, xin2021neurolkh}}, both of which have well-studied practical variants~\mbox{\cite{braekers2016vehicle, gunawan2016orienteering}}. Practical use cases for parameter-constrained complexity scaling laws are more challenging to identify given that training until convergence is usually compute-inefficient. One use case would be predicting performance for applications mainly constrained by model size, such as edge applications requiring fast inference with limited hardware acceleration. Another potential use case is algorithm benchmarking, where we can estimate the limit of performance for several algorithms across a range of problem complexities without training models on larger problem scales. Such benchmarks can also provide useful insights like sample efficiency tradeoffs with respect to problem complexity. Our single-seed node scaling results for RL and SFT demonstrate that SFT reaches intermediate suboptimality values faster at the scales evaluated (Figure~\mbox{\ref{fig:complexitybreaktemporal}} in Appendix~\mbox{\ref{apx:trendbreak}}). Future work could use multiple-seed training to quantify this advantage with high statistical confidence.

\paragraph{Finite-compute complexity scaling:} While this paper focuses on bottleneck from model capacity, we could have instead chosen to limit compute budget or dataset size (with early stopping) and observe
the patterns that emerge with increased problem complexity, if any. Compute-constrained and data-constrained complexity scaling laws would have many practical use cases. For example, Jones~\cite{jones2021scaling} shows that we can predict model performance as a bivariate function of train-time compute and test-time compute for a single Hex board size. If this neural scaling law can be generalized over problem scale, one could estimate for complex tasks the tradeoff between train-time and test-time compute using only cheap experiments.

\begin{ack}
We thank Charles Lovering for his feedback on delivery and clarity. We thank Joseph Weissman for reviewing portions of our proofs. This work was performed without third-party support.
\end{ack}

\newpage
\addcontentsline{toc}{section}{References}
\bibliographystyle{unsrtnat}  
\bibliography{references}

\newpage

\tableofcontents

\newpage
\appendix
\section*{Appendix}

\section{Scaling law fit constants}
\label{apx:fits}

We provide the fit constants for scaling laws discussed in the main paper. Remaining fits can be found in the project repository at \url{https://github.com/lowellw6/complexity-scaling-laws}. For example, our repository provides fits of the scaling laws for loss discussed in Appendix \ref{apx:loss}.

RL and SFT scaling laws may have non-trivial imprecision due to training with a single seed per scale. This should be considered when extrapolating, or when comparing values between RL and SFT such as \(\alpha\) in node scaling and optimal model size fits. To mitigate this concern, we performed jackknife (leave-one-out) resampling of RL and SFT fits and provide growth rate statistics in Table~\mbox{\ref{tab:hoo}}. We provide individual leave-one-out (LOO) fits in the project repository. While LOO distributions~are distinct from those obtained using multiple seeds, they quantify sensitivity to individual training runs.

Optimal model size fits are relatively sensitive due to the frontier overlap between adjacent model sizes. We observe higher variance and the largest SFT LOO \mbox{\(\alpha\)} exceeds the smallest RL LOO \mbox{\(\alpha\)} (despite the full-data RL \mbox{\(\alpha\)} being larger than the full-data SFT \mbox{\(\alpha\))}. Node-scaling fits are relatively sensitive to the suboptimalities obtained at the largest node scales. For RL, we obtain \mbox{\(\alpha\)} values as large as \mbox{\(1.95\)} when fixing the \mbox{\(\gamma\)} offset at its full-data value and \mbox{\(2.22\)} when fitting \mbox{\(\gamma\)}. Demonstrated by the large positive skewness, these larger \mbox{\(\alpha\)} fits occur at larger LOO node scales. However, fitting \mbox{\(\gamma\)} can shift the offset more than \mbox{\(3\)} nodes and we know near optimality is achieved at \mbox{\(5\)} nodes, so fixed-\mbox{\(\gamma\)} LOO variance is likely closer to multi-seed variance. Remaining LOO fits are quite stable, and we found that \mbox{\(\beta\)} asymptote fits for dimension scaling experiments are nearly constant (roughly \mbox{\(\pm0.001\)}).

\begin{table}[h]
    \caption{Neural scaling law fits.}
    \centering
    \begin{tabular}{ccccc}
        \toprule
        Scaling law & Form & Algorithm & \(\alpha\) & \(k\)  \\
        \midrule
        \multirow{2}{*}{Parameters (\(N\))} & \multirow{2}{*}{\(s=\left(\frac{k}{N}\right)^{\alpha}\)} & RL & \(0.582\) & \(4.42 \times 10^2\)  \\
        & & SFT & \(0.712\) & \(1.38 \times 10^3\)  \\
        \midrule
        \multirow{2}{*}{Compute (\(C_{\min}\))} & \multirow{2}{*}{\(s=\left(\frac{k}{C_{\min}}\right)^{\alpha}\)} & RL & \(0.439\) & \(6.62 \times 10^{-9}\)  \\
        & & SFT & \(0.555\) & \(6.78 \times 10^{-9}\)  \\
        \midrule
        \multirow{2}{*}{Optimal model size} & \multirow{2}{*}{\(N=\left(\frac{C}{k}\right)^{\alpha}\)} & RL & \(0.816\) & \(1.75 \times 10^{-12}\)  \\
        & & SFT & \(0.658\) & \(1.98 \times 10^{-14}\)  \\
        \bottomrule
    \end{tabular}
    \label{tab:neuralfits}
\end{table}

\begin{table}[h]
    \caption{TSP node (\(n\)) scaling law fits.}
    \centering
    \begin{tabular}{cccccc}
        \toprule
        Form & Algorithm & \(M\) & \(\alpha\) & \(\gamma\) & \(k\)  \\
        \midrule
        \multirow{7}{*}{\(s=\left(\frac{n-\gamma}{k}\right)^{\alpha}\)} & RL & -- & \(1.86\) & \(4.99\) & \(1.91 \times 10^2\)  \\
        & SFT & -- & \(1.69\) & \(4.99\) & \(2.24 \times 10^2\)  \\
        & 2-opt & \(25\) & \(2.09\) & \(22.3\) & \(18.6\)  \\
        & 2-opt & \(20\) & \(1.82\) & \(21.0\) & \(14.2\)  \\
        & 2-opt & \(15\) & \(1.67\) & \(18.1\) & \(11.1\)  \\
        & 2-opt & \(10\) & \(1.49\) & \(14.8\) & \(8.05\)  \\
        & 2-opt & \(5\) & \(1.27\) & \(10.8\) & \(4.94\)  \\
        \bottomrule
    \end{tabular}
    \label{tab:nodefits}
\end{table}

\begin{table}[h]
    \caption{TSP spatial dimension (\(d\)) scaling law fits. RL fits use pure exponential decay (\(\phi=1\)).}
    \centering
    \begin{tabular}{ccccccc}
        \toprule
        Form & Algorithm & \(n\) & \(\beta\) & \(\psi\) & \(\phi\) & \(k\)  \\
        \midrule
        \multirow{4}{*}{\(s=\beta - \left(\frac{k}{\psi^{d^{\phi}}}\right)\)} & RL & \(10\) & \(5.33 \times 10^{-2}\) & \(1.18\) & \(1.00\) & \(7.24 \times 10^{-2}\)  \\
        & RL & \(20\) & \(0.209\) & \(1.66\) & \(1.00\) & \(0.557\)  \\
        & 2-opt & \(10\) & \(5.70 \times 10^{-2}\) & \(66.8\) & \(0.215\) & \(5.22\)  \\
        & 2-opt & \(20\) & \(0.254\) & \(98.6\) & \(0.251\) & \(38.0\)  \\

        \bottomrule
    \end{tabular}
    \label{tab:dimfits}
\end{table}

\FloatBarrier

\begin{table}[h]
    \captionsetup{justification=centering}
    \caption{Leave-one-out (LOO) statistics for RL and SFT fits of exponent \mbox{\(\alpha\)} and base \mbox{\(\psi\)}. \\ LOO indices (idx) are in ascending scale order.}
    \centering
    \begin{tabular}{cccccrcc}
        \toprule
        Scaling law & Algorithm & Fit & \(\mu\) & \(\sigma\) & Skew & Min (idx) & Max (idx) \\
        \midrule
        \multirow{2}{*}{Parameters (\(N\))} & RL & \(\alpha\) & \(0.579\) & \(0.013\) & \(-1.9\) & \(0.545\,(0)\) & \(0.593\,(4)\)  \\
        & SFT & \(\alpha\) & \(0.718\) & \(0.015\) & \(0.87\) & \(0.701\,(2)\) & \(0.745\,(0)\) \\
        \midrule
        \multirow{2}{*}{Compute (\(C_{\min}\))} & RL & \(\alpha\) & \(0.440\) & \(0.011\) & \(0.83\) & \(0.423\,(5)\) & \(0.463\,(0)\)  \\
        & SFT & \(\alpha\) & \(0.554\) & \(0.011\) & \(-0.28\) & \(0.536\,(2)\) & \(0.570\,(0)\) \\
        \midrule
        \multirow{2}{*}{Optimal model size} & RL & \(\alpha\) & \(0.811\) & \(0.064\) & \(-0.47\) & \(0.690\,(7)\) & \(0.903\,(5)\)  \\
        & SFT & \(\alpha\) & \(0.655\) & \(0.071\) & \(-0.22\) & \(0.556\,(4)\) & \(0.734\,(2)\) \\
        \midrule        
        \multirow{2}{*}{Nodes (\(n\)) - fix \(\gamma\)} & RL & \(\alpha\) & \(1.87\) & \(0.032\) & \(1.4\) & \(1.83\,(6)\) & \(1.95\,(8)\) \\
        & SFT & \(\alpha\) & \(1.71\) & \(0.059\) & \(1.4\) & \(1.65\,(3)\) & \(1.83\,(5)\) \\
        \midrule
        \multirow{2}{*}{Nodes (\(n\)) - fit \(\gamma\)\:} & RL & \(\alpha\) & \(1.90\) & \(0.116\) & \(2.2\) & \(1.81\,(0)\) & \(2.22\,(8)\) \\
        & SFT & \(\alpha\) & \(1.69\) & \(0.134\) & \(0.46\) & \(1.48\,(0)\) & \(1.94\,(5)\) \\
        \midrule
        \multirow{2}{*}{Dimensions (\(d\)) \; \begin{tabular}{@{}c@{}} 10\,\(n\) \\ 20\,\(n\) \end{tabular}} & RL & \(\psi\) & \(1.18\) & \(0.007\) & \(-1.0\) & \(1.16\,(0)\) & \(1.19\,(5)\) \\
         & RL & \(\psi\) & \(1.68\) & \(0.023\) & \(-0.46\) & \(1.62\,(2)\) & \(1.71\,(0)\) \\
        \bottomrule
    \end{tabular}
    \label{tab:hoo}
\end{table}

\begin{table}[h]
    \captionsetup{justification=centering}
    \caption{Performance bounds obtained for TSP node and spatial dimension scaling. Random tour length w.r.t. nodes is derived (Lemma \ref{proof:expn}). Otherwise, these fits should \textit{not} be extrapolated.}
    \centering
    \begin{tabular}{cccccrc}
        \toprule
        Scaling variable & Form & Algorithm & \(n\) & \(\alpha\) & \(\gamma \quad\) & \(k\)  \\
        \midrule
        \multirow{2}{*}{Nodes (\(n\))} & \multirow{2}{*}{\(\mu_{\text{cost}}=\left(\frac{n-\gamma}{k}\right)^{\alpha}\)} & Optimal & -- & \(0.433\) & \(-0.215\) & \(0.906\)  \\
        & & Random & -- & \(1.000\) & \(0.000\) & \(1.918\)  \\
        \midrule
        \multirow{4}{*}{Dimensions (\(d\))} & \multirow{4}{*}{\(\mu_{\text{cost}}=\left(\frac{d-\gamma}{k}\right)^{\alpha}\)} & Optimal & \(10\) & \(0.593\) & \(0.981\) & \(0.175\)  \\
        & & Random & \(10\) & \(0.501\) & \(0.368\) & \(6.04 \times 10^{-2}\)  \\
        & & Optimal & \(20\) & \(0.656\) & \(1.24\) & \(0.102\)  \\
        & & Random & \(20\) & \(0.500\) & \(0.369\) & \(1.50 \times 10^{-2}\)  \\
        \bottomrule
    \end{tabular}
    \label{tab:boundfits}
\end{table}

\newpage
In Table \ref{tab:boundfits}, for random tour length and optimal tour length, we provide the scaling trends that we obtain for the tested TSP node and spatial dimension domains. Random tour length w.r.t. number of nodes is derived (Lemma \ref{proof:expn}). Other constants listed are empirical fits that may be useful for interpolation. For example, mean optimal tour length w.r.t. nodes produces errors less than \(5\times10^{-3}\) within the tested domain of 5 to 50 nodes. However, these fits should \textit{not} be extrapolated to much larger scales because several of these fits provably break down. For mean optimal tour length, \(\alpha\to0.5\) as \(n\to\infty\) \cite{beardwood1959shortest}, where our empirical fit is \(\alpha=0.433\). For mean random tour length w.r.t. spatial dimensions, \(\alpha\) fits are close to the true limiting behavior \(\alpha=0.5\) (Theorem \ref{proof:limd}), but error eventually accumulates due to inexact proportionality constant fits.

\FloatBarrier
\section{Trend breakdown and assessment of learning convergence}
\label{apx:trendbreak}
This section details larger-scale experiments that did not converge to trend. We also elaborate on relevant details of our training method. We show models which visibly failed to converge using this method and we identify boundary cases where convergence is more difficult to assess. These models underperform our scaling law predictions, evaluating to larger suboptimality values. If any model were to outperform our scaling law predictions, trend breakdown could not be attributed to insufficient training.

A significant portion of neural scaling laws research is performed by organizations with access to abundant computational resources. Experiments that initially fail to converge (if any) can be rapidly iterated. Researchers with more computational constraints often do include results for trend-breaking experiments. However, these results are sometimes mentioned in passing without evaluating model convergence. Besides promoting transparency, we provide this appendix to document the behavior of scaling law breakdown under compute budget constraints.

\begin{figure}[h]
    \centering
    \includegraphics[clip, trim=0.1in 0.1in 0.1in 0.1in, width=5.5in]{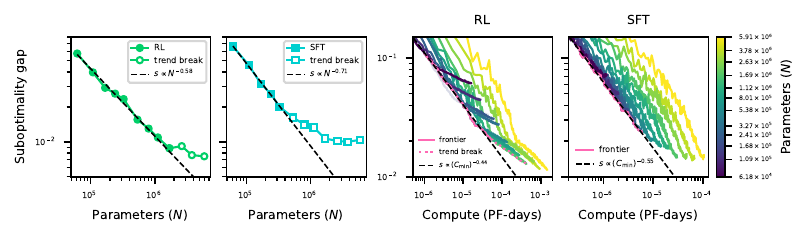}
    \caption{Suboptimality w.r.t. parameter and compute scaling, including experiments which did not converge to trend. \textbf{Left pair:} Parameter scaling experiments evaluated at the end of training. Unfilled markers indicate models not shown in Figure~\ref{fig:sjoint} that are excluded from power law fits. \textbf{Right~pair:} Suboptimality evaluated throughout training w.r.t. compute. Models excluded from parameter scaling fits fail to reach the compute-efficient frontier with our training settings.}
    \label{fig:sjointbreak}
\end{figure}

\subsection{Training details}
Without data constraints, RL models trained for one million gradient updates with a three-stage learning rate schedule: a brief linear warm-up, then cosine decay over early convergence, then a slow linear decay to zero. Cosine decay terminates at a learning rate of \(10^{-5}\) at update number 170,000. This termination update was informed by hyperparameter optimization results that we detail in Appendix~\ref{apx:hpo}. For the remainder of our training budget, 830,000 updates, it was a design choice to use a slow linear learning rate decay. For models which nearly converge within the cosine decay window, this approach reliably extracts remaining marginal improvements. However, late-stage training consumes most of the training budget. Models which insufficiently converge within the cosine decay window make less progress than they would if the learning rate was instead reduced more gradually. We opted for this trade-off to prioritize ensuring convergence for experiments at smaller scales, providing more precise results for the scaling laws we present in the main paper.

For SFT models, data constraints restricted training duration given we stop after one epoch to avoid potential performance bottlenecks from dataset size. For these experiments, after warm-up, the learning rate cosine decays to zero over the remainder of the epoch, which yields 73,143 updates with our dataset and batch size settings. Despite this abbreviation in training, SFT's strong supervised learning signal results in sufficient convergence for several smaller model scales and problem scales.

\subsection{Neural scaling assessment}
Figure \ref{fig:sjointbreak} shows the complete set of parameter and compute scaling experiments we evaluated. This includes the fitted results shown in Figure~\ref{fig:sjoint} along with results from larger models that failed to converge. Parameter scaling evaluations drift right of trend and level off. Temporal compute scaling trends demonstrate that these models failed to converge with our training settings. Learning curves for larger SFT models maintain steep descents throughout training. Curves for larger RL models level off only slightly, which is attributable to the very small learning rates used in late-stage training.

\paragraph{168,000 parameter outlier:} We also exclude the roughly 168,000 parameter model (shown in lower opacity) when fitting the RL compute-efficient frontier. This outlier is unrelated to convergence. However, fitting with this model separates the frontier from several other models that also converge. Scaling laws predict the expectation of model performance over random seed variables like parameter initializations and training batch sampling, and this outlier's random seed appears to perform unusually well. Averaging evaluations over several seeds would likely address this discrepancy.

\begin{figure}[h]
    \centering
    \includegraphics[clip, trim=0.1in 0.1in 0.1in 0.1in, width=5.5in]{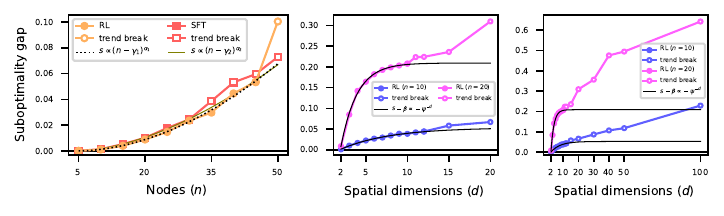}
    \caption{Suboptimality w.r.t. TSP node and spatial dimension scaling, including experiments which did not converge to trend. Unfilled markers indicate models not shown in Figure~\ref{fig:complexity} that are excluded from scaling law fits. \textbf{Left:} Node scaling experiments evaluated at the end of training. SFT's power growth fit is shown in a different color only for clarity. \textbf{Right~pair:} Spatial dimension scaling experiments evaluated at the end of training. We show early trend breakdown (center) along with the full domain up to 100 dimensions (right).}
    \label{fig:complexitybreak}
\end{figure}

\subsection{Complexity scaling assessment}
\begin{figure}
    \centering
    \includegraphics[clip, trim=0.1in 0.1in 0.1in 0.1in, width=5.5in]{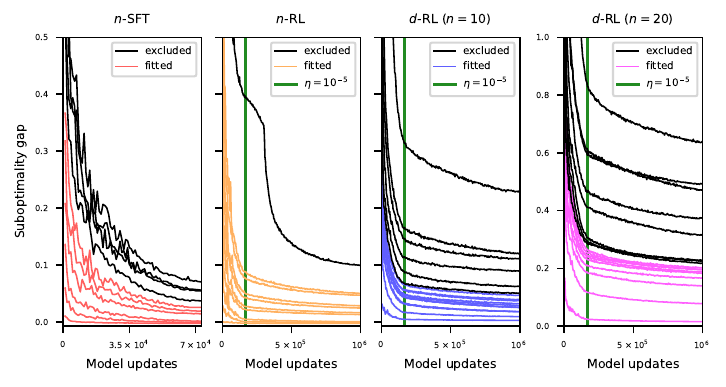}
    \caption{Suboptimality w.r.t. TSP node (\(n\)) and spatial dimension (\(d\)) scaling evaluated throughout training. For RL experiments, the end of cosine decay is marked with a green vertical line. Subsequent updates use learning rates below \(10^{-5}\), which impedes the convergence of models training on larger problem scales. Note that we use smaller 12,800 sample datasets to evaluate each learning curve. Each dataset is distinct since we scale the problem, so small differences between adjacent scales may not be statistically significant. The reduced sample size also (falsely) produces negative suboptimality values for 2D TSP with 5-node and 10-node scales where model performance becomes near-optimal.}
    \label{fig:complexitybreaktemporal}
\end{figure}

Figure \ref{fig:complexitybreak} shows the full set of TSP node and spatial dimension scaling experiments we evaluated. For node scaling results, excluded SFT experiments only slightly break trend, whereas the excluded 50-node RL experiment sharply breaks trend. Figure~\ref{fig:complexitybreaktemporal} explains this behavior with suboptimality evaluations throughout training. Learning curves for excluded SFT experiments are more divergent and maintain steeper descents until learning rate decay impedes progress. And the excluded 50-node RL experiment performs particularly poorly in early training during the cosine decay period. Subsequent training makes significant progress, so this model may have initially approached a poor local optimum in loss landscape.

For spatial dimension scaling results, excluded experiments at scales beyond 20 dimensions fail to sufficiently converge in early training and maintain steeper descents in later training (right pair of Figure~\ref{fig:complexitybreaktemporal}). However, excluded experiments between 11 and 15 dimensions are not as distinguishable from their fitted counterparts, especially with non-trivial imprecision from using smaller evaluation datasets that vary for each scale. These experiments correspond to the lowest black curve for the 10-node plot (15 dimensions) and the lowest three black curves for the 20-node plot (11, 12, and 15 dimensions). For these boundary cases, we solely rely on evidence from adjacent scales to surmise that further training can produce trend alignment.

\FloatBarrier
\section{Loss behavior}
\label{apx:loss}

For generative modeling, the majority of research on neural scaling laws studies cross-entropy loss~\cite{kaplan2020scalinglawsneurallanguage, henighan2020scaling, hoffmann2022training}. However, in general, we do not expect loss to trend for RL algorithms~\cite{hilton2023scaling}. Further, the scaling behavior of RL loss is often irrelevant to task performance, whereas cross-entropy loss in language modeling, for example, is directly relevant to task performance. This detachment from performance makes scaling laws for RL loss less interpretable.

Even so, we observe several loss behaviors worth documenting. We detail these behaviors below and summarize them here. First, separate loss components can concurrently trend and not trend even when these components have interdependent learning objectives. Second, variance of loss can reveal an underlying trend when mean loss does not trend. Third, when scaling problem complexity, token-level cross-entropy loss may not correlate with task performance.

\subsection{Relevant experiment details}
SFT models were trained to predict optimal tour length with a critic head alongside the primary objective of minimizing negative log-likelihood (NLL) of optimal next-node prediction. This setup maintains identical architectures (and equal parameter counts) between SFT and RL experiments, isolating differences to the policy learning objective.

Loss evaluations use the same datasets as corresponding suboptimality evaluations. We evaluate the same scales used for the main paper as we found that loss trend breakdown closely aligns with suboptimality trend breakdown when using our training method (with limited compute). Suboptimality trend breakdown is discussed in Appendix~\ref{apx:trendbreak}.

\begin{figure}[b]
    \centering
    \includegraphics[clip, trim=0.1in 0.1in 0.1in 0.1in, width=5.5in]{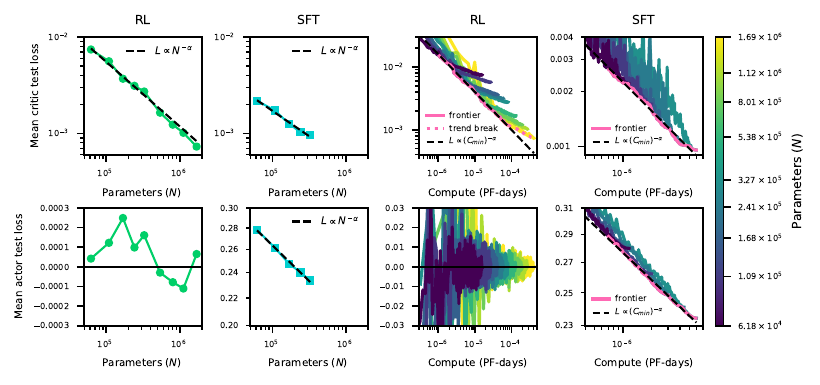}
    \caption{Loss w.r.t. parameter and compute scaling. Note that SFT's compute axes (right column) have been stretched for easier viewing, and the loss axes for those subplots do not share scale with the RL compute scaling subplots to their left (unlike Figure~\ref{fig:sjoint}). \textbf{Top:} Self-supervised critic loss smoothly power decays when predicting on-policy tour length (RL) or predicting optimal tour length (SFT). \textbf{Bottom:} Supervised NLL actor loss smoothly power decays but PPO actor loss has no mean trend other than being zero-centered w.r.t. the non-stationary critic baseline.}
    \label{fig:ljoint}
\end{figure}

\begin{figure}[h]
    \centering
    \includegraphics[clip, trim=0.1in 0.1in 0.1in 0.1in, width=5.5in]{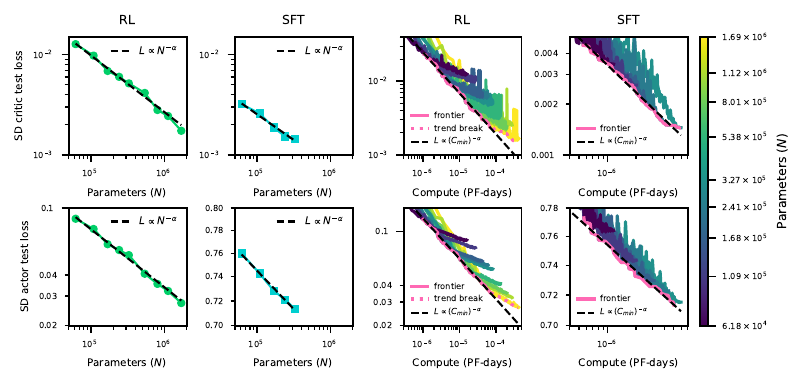}
    \caption{Standard deviation (SD) of loss w.r.t. parameter and compute scaling. Subplot layout matches that used for Figure~\ref{fig:ljoint}. The variance of each loss component exhibits smooth power decay, even for the PPO actor loss which does not produce scaling laws for mean loss.}
    \label{fig:ljointvar}
\end{figure}

\subsection{Neural scaling loss behaviors}
Figure \ref{fig:ljoint} shows model test loss w.r.t. parameter and compute scaling. Supervised NLL loss power decays as expected from previous neural scaling laws work. PPO's clipped surrogate objective~\cite{schulman2017proximalpolicyoptimizationalgorithms} (actor loss) has no obvious trend other than being zero-centered, also as expected. However, PPO's critic loss does smoothly power decay. The existence of this scaling law is more surprising given the value target distribution is non-stationary (following on-policy tour length) and the mean actor loss does not trend.

The behavior PPO actor loss \textit{variance} may partially explain why scaling laws emerge for mean PPO critic loss. Figure~\ref{fig:ljointvar} shows that variance of loss is also predictable when scaling parameters or compute. This observation is perhaps intuitive for loss components where the mean trends. However, we obtain similarly precise scaling laws for the variance of PPO actor loss, revealing that an underlying predictable nature persists despite the absence of a predictable mean. The existence of scaling laws for mean critic loss may depend on this predictable actor loss behavior, and vice versa, given the strong interdependence between the learning objectives.

Lastly, a subtle additional finding from these results is that scaling laws for loss appear more robust against outliers when compared to their suboptimality counterparts. For suboptimality trends in Figure~\ref{fig:sjoint}, including the (roughly) 168,000 parameter model in the RL compute-efficient frontier fit noticeably skews the result. In contrast, while this outlier is still visible in the RL compute-efficient frontier for critic loss (Figure~\ref{fig:ljoint} top, third column), excluding it hardly alters the obtained fit.

\subsection{Complexity scaling loss behaviors}
\label{losscomplexityscaling}
\begin{figure}
    \centering
    \includegraphics[clip, trim=0.1in 0.1in 0.1in 0.09in, width=5.5in]{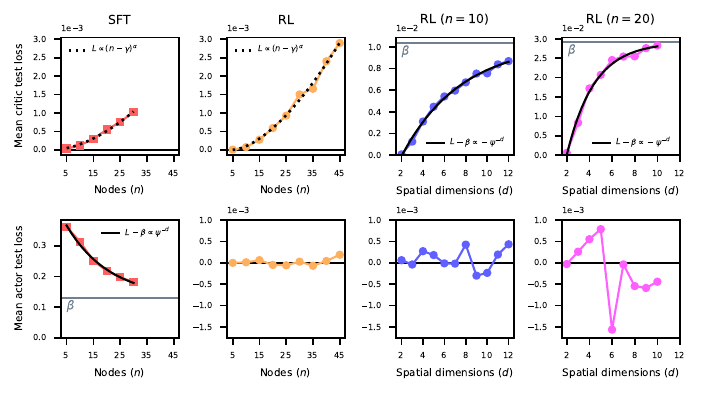}
    \caption{Loss w.r.t. TSP node and spatial dimension scaling. \textbf{Top:} Self-supervised critic loss adheres to the corresponding scaling law form observed for suboptimality (Figure~\ref{fig:complexity}). \textbf{Bottom:} PPO actor loss only follows a zero-centered trend, as it does for neural scaling experiments (Figure~\ref{fig:ljoint}). Supervised NLL loss adheres to exponential decay when scaling nodes, deviating from the power growth pattern observed for suboptimality and critic loss. This trend implies that next-node prediction improves with node scale for SFT models, at least over smaller scales. However, this trend does \textit{not} imply that full-tour prediction improves because the number of nodes predicted per tour increases.}
    \label{fig:lcomplexity}
\end{figure}

Figure \ref{fig:lcomplexity} shows model test loss w.r.t. TSP node and spatial dimension scaling. Aligning with observations for parameter and compute scaling, self-supervised critic loss components mirror the scaling law forms of their suboptimality counterparts, and PPO actor loss is zero-centered. However, supervised NLL loss w.r.t. node scale deviates from the power growth trend that SFT suboptimality and optimal tour length prediction adhere to. Instead, NLL loss exponentially decays with increased node scale over the tested domain.\footnote{We also tested power decay but this form converges too slowly and fits a negative \(\beta\) asymptote (NLL is strictly non-negative).} Hence, as suboptimality accumulates with increasing solution space complexity, next-node prediction improves. 

One potential explanation is that next-node prediction benefits from the increasing negative supervision: each label for next-node selection provides one positive example and (\(n-1\)) negative examples. If we were to assume this trend extrapolates, then next-node prediction NLL converges as suboptimality diverges. If so, the joint NLL of full-tour prediction may still diverge given the number node predictions per tour increases, although we did not evaluate this metric. Alternatively, the trend observed for node-level NLL loss may break down shortly after the tested scales. In any case, node-level cross-entropy loss does not correlate with task performance. This complication is avoided when scaling laws are obtained for metrics that correlate with task performance by definition (for example, suboptimality).

\begin{figure}
    \centering
    \includegraphics[clip, trim=0.1in 0.1in 0.1in 0.09in, width=5.5in]{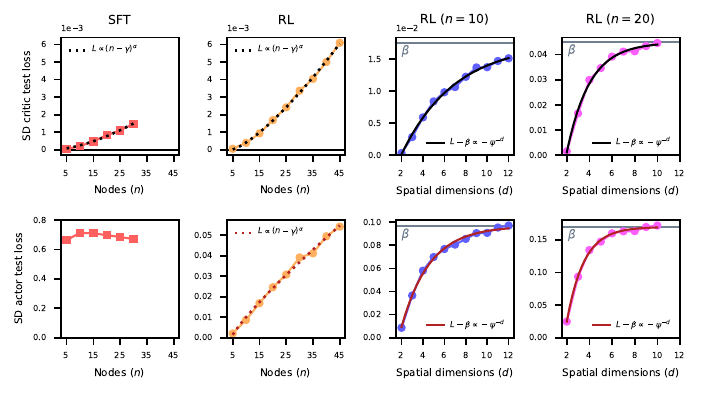}
    \caption{Standard deviation (SD) of loss w.r.t. TSP node and spatial dimension scaling. Subplot layout matches that used for Figure~\ref{fig:lcomplexity}. \textbf{Top:} Standard deviation of critic loss adheres to the corresponding scaling law form of mean critic loss (and suboptimality). \textbf{Bottom:} Standard deviation of PPO actor loss roughly aligns with the corresponding suboptimality and critic loss trends, but fit quality is markedly worse (illustrated in red). Standard deviation of supervised NLL loss w.r.t. node scale does not follow a consistent trend (bottom left).}
    \label{fig:lcomplexityvar}
\end{figure}

We find that standard deviation (SD) of critic loss follows the corresponding scaling law form of suboptimality and mean critic loss, as shown in Figure~\ref{fig:lcomplexityvar}. However, scaling behaviors are not as predictable for SD of PPO actor loss or supervised NLL. For the latter, node scale has no obvious correlation with SD. For the PPO actor loss, SD trends resemble their critic loss and suboptimality counterparts but fits are clearly inferior. We observe contiguous residuals of the same sign, and for spatial dimension scaling experiments exponential decay converges too fast, fitting \(\beta\) values below the SD values obtained at the largest scales. Compared to neural scaling results in Figure~\ref{fig:ljointvar}, predicting SD of loss appears to be more nuanced when scaling complexity. However, note that SD and variance would not share the dimension scaling law form due to the additive \(\beta\) term. Actor loss variance may produce a closer fit, but we did not evaluate these trends.

\FloatBarrier
\section{Local search supplement}
\label{apx:landscape}
Section~\ref{landscapeproperties} details the 2-opt cost landscape properties referenced in Sections \ref{repscale} and \ref{searchreproduce}. Section~\ref{2excresults} provides complexity scaling results for 2-exchange search, which uses a different definition of solution adjacency than 2-opt search.

\subsection{Properties of the 2-opt cost landscape}
\label{landscapeproperties}
For each complexity scaling scaling experiment, we measure three fundamental properties of the 2-opt cost landscape in expectation: the number of local optima, the size of a local optimum's basin of attraction, and the relative size of the global optimum's basin of attraction. Results in Figure~\ref{fig:2opt_landscape} suggest that these properties converge when scaling TSP spatial dimensions. In contrast, these properties diverge when scaling the number of TSP nodes. 

\subsubsection{Evaluation method}
To measure these properties, we perform numerous search descents (\(k\)) for each TSP problem evaluated. Each descent starts at a random tour and terminates when arriving at a locally optimal tour (w.r.t. one additional 2-opt move). We then count the number of unique local optima found, count the number of search moves required to reach them (basin of attraction size), and evaluate the fraction of descents that arrived at the best-found local optima (relative size of the global optimum's basin of attraction).

Estimating the number of local optima requires adequate coverage of the landscape, as does estimating the visitation rate of the global optimum given we approximate the global optimum using the best-found local optima (BLO). For these estimations, we perform 100,000 descents per problem (large-\(k\)) while sampling small batches for each scale: 2048 problems for 10-node dimension scaling, 512 problems for 20-node dimension scaling, and 32 problems for node scaling.

Evaluating mean search moves does not require adequate coverage of the landscape, so we use fewer descents per problem and sample larger problem batches (small-\(k\), large-batch). For dimension scaling, we reuse the approximately optimal datasets detailed in Appendix~\ref{apx:proxyval}. For node scaling, we use 100 descents per problem and sample 64,000 problems. We also show evaluations for local optima count and BLO visitation rate using these datasets for comparison (blue triangles in Figure~\ref{fig:2opt_landscape}).

\subsubsection{Property convergence as \texorpdfstring{\(d\to\infty\)}{dinfty}}
\begin{figure}
    \centering
    \includegraphics[clip, trim=0.1in 0.1in 0.1in 0.0in, width=5.2in]{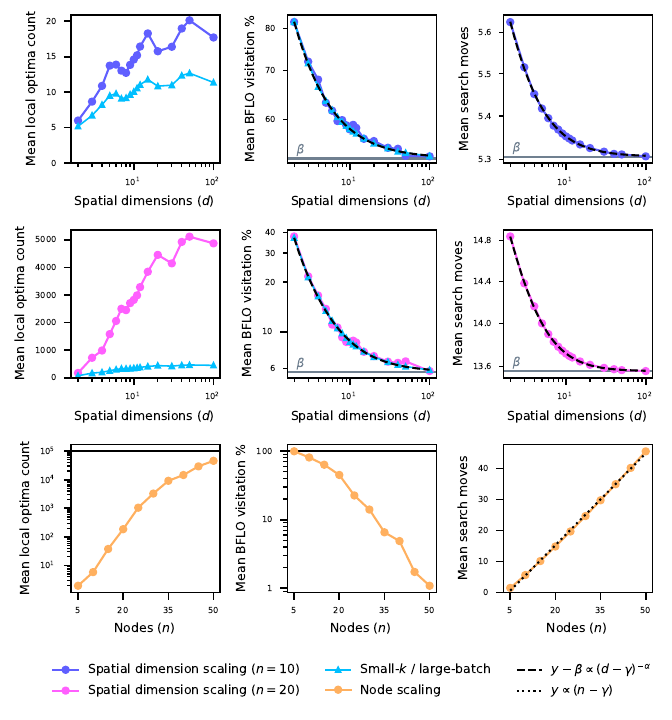}
    \caption{2-opt cost landscape properties w.r.t. TSP node and spatial dimension scaling. From top to bottom: 10-node spatial dimension scaling, 20-node spatial dimension scaling, and 2-dimensional node scaling. From left to right: mean local optima count, mean frequency that search arrives at the best-found local optimum (BLO), and mean search depth before arriving at a local optimum. \textbf{Dimension scaling:} The number of local optima exhibits non-monotonic growth that is upper bounded by the fixed size of the solution space. BLO visitation rate and search depth adhere to power decay. We found that including a \(\gamma\) scale offset improves fit quality, but excluding \(\gamma\) still produces reasonable fits. From left to right axis scalings are: [semi-log-x,~log-log,~log-log]. \textbf{Node scaling:} Due to the expanding solution space, the number of local optima diverges as does the search depth required to reach one. BLO visitation rate decays roughly exponentially. Note that local optima count is underestimated by 100,000 descents per problem~\cite{tayarani2016analysis}. From left to right scalings are: [semi-log-y,~semi-log-y,~linear].}
    \label{fig:2opt_landscape}
\end{figure}

Figure~\ref{fig:2opt_landscape} shows that expected search depth and BLO visitation rate both power decay when increasing spatial dimensions. The former implies that basin of attraction size converges in expectation. Because each 2-opt search move evaluates \(O(n^2)\) adjacent solutions, and TSP node scale remains constant, convergent search depth implies convergence of the number of solutions evaluated per descent. Alongside the convergence of the BLO visitation rate, these observations suggest that, with many-descent local search, finding the optimal solution approaches a stationary computational complexity apart from the \(O(d)\) complexity of evaluating each visited solution's cost.

Unlike search depth and BLO visitation rate, the number of local optima has no predictable convergent trend and instead grows non-monotonically. This is not an artifact of sample size as we observe the same behavior for small-\(k\), large-batch evaluations (with a significant underestimation bias). Growth visibly slows by 100 dimensions, and growth is upper bounded by the fixed solution space size, so convergence is plausible but not directly implied by these results.

Lastly, these properties are closely related to landscape ruggedness measures like autocorrelation~\cite{angel2000classification, angel2001landscape}. Because the expected range of cost values also converges (Theorem~\ref{proof:constspan}), the previous findings may imply autocorrelation converges as \mbox{\(d\to\infty\)}. However, we did not directly evaluate this measure.

\subsubsection{Property divergence as \texorpdfstring{\(n\to\infty\)}{ninfty}}
Node scaling produces none of the convergent complexity behaviors discussed above. Cost landscape analysis over solution space scaling is a well-studied topic in TSP and combinatorial optimization \cite{tayarani2016analysis} (Section 1.1 in the cited work provides a useful overview). Our result is provided for convenience.

When scaling nodes, the number of local optima rapidly exceeds what we can accurately estimate with 100,000 descents per problem. For 3-opt search moves, Tayarani et al.~\cite{tayarani2016analysis} show that this growth appears to be \(O(e^{n\ln(n)})\). 2-opt's search depth grows almost linearly and 2-opt's BLO visitation rate decays roughly exponentially. Both observations align with 3-opt findings~\cite{tayarani2016analysis}. Note that BLO visitation rate converging toward zero does not reflect a steady state property of the cost landscape: we are measuring the relative size of the global optimum's basin of attraction as the absolute size of the landscape diverges.

\subsection{2-exchange search results}
\label{2excresults}
In this paper, we define "2-exchange" as the naive search move that swaps two nodes in the tour sequence. We are not the first to use this definition~\cite{hernando2011study}; however, the term is sometimes used to refer to 2-opt in informal sources. 2-opt is the distinct search move that inverts a subtour~\cite{croes1958method}.

\begin{figure}[h]
    \centering
    \includegraphics[clip, trim=0.1in 0.12in 0.1in 0.1in, width=5.5in]{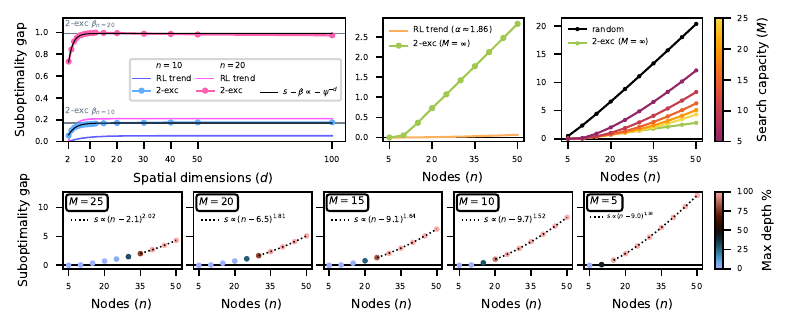}
    \caption{2-exchange local search suboptimality over problem complexity scaling. When scaling nodes and constraining search depth, suboptimality follows superlinear power growth after saturating at 100\% max depth (top right and bottom), aligning with 2-opt behavior and the scaling form of parameter-constrained models. Like RL fits, 2-exchange fits do not benefit from fitting the subexponential constant \(\phi\) when scaling spatial dimensions. However, 2-exchange dimension scaling fits are slightly inferior to their 2-opt counterparts, which may be attributable to 2-exchange exhibiting several of the random-suboptimality behaviors shown in Figure~\ref{fig:bounds}. We observe a slight decrease near convergence when scaling dimensions for the 20-node experiment, and unconstrained 2-exchange search produces linear suboptimality growth when scaling nodes (beyond 10 nodes).}
    \label{fig:jointsearch_2exc}
\end{figure}

Figure~\ref{fig:jointsearch_2exc} shows that 2-exchange reproduces 2-opt's alignment with the complexity scaling law forms of parameter-constrained models (Figure~\ref{fig:jointsearch}). However, both moves modify only a few edges and both adjacency sets scale \(O(n^2)\), so this result demonstrates limited generalization. Analyzing the conditions in which this alignment generalizes for \(k\)-opt moves~\cite{lin1973effective} is one potential direction for future work.

\FloatBarrier
\newpage
\section{Optimal solution approximation in higher dimensions}
\label{apx:proxyval}
For node scaling evaluations we leverage the Concorde-generated optimal solutions introduced in Section~\ref{optsolgen}. But evaluating suboptimality in higher dimensions requires a different approach since the PyConcorde software stack expects 2D TSP. To the best of our knowledge, applying Concorde to higher-dimensional TSP has not been explored. This appears to be only an implementation constraint, though, and not a limitation of the Concorde algorithm itself since it uses the cutting-plane method \cite{applegate1998solution} and additional spatial dimensions only modify the edge cost calculation. Regardless, we opted to closely approximate optimality for convenience.

\subsection{Approximation method}
Near-optimal solution tours are generated via mass repetition of local search. For a single TSP instance, \(k\) random starting tours are sampled, and for each starting tour a distinct descent of local search is performed. Among the \(k\) generated locally optimal tours, the tour with lowest cost is selected as a \textit{surrogate} for the globally optimal tour. This full process is repeated \(B\) times using a large batch of sampled TSP instances to acquire near-optimal datasets at a desired problem scale.

We use 2-opt moves when generating all surrogate optimal datasets. There are more sophisticated local search algorithms that usually find lower cost solutions for TSP \cite{blazinskas2011combining, lin1973effective}, though often with increased search time or compute. But we find 2-opt surrogate optimality to be sufficiently accurate at the needed problem scales despite its simplicity.

\begin{figure}[b]
    \centering
    \includegraphics[clip, trim=0in 0.1in 0in 0in, width=5.0in]{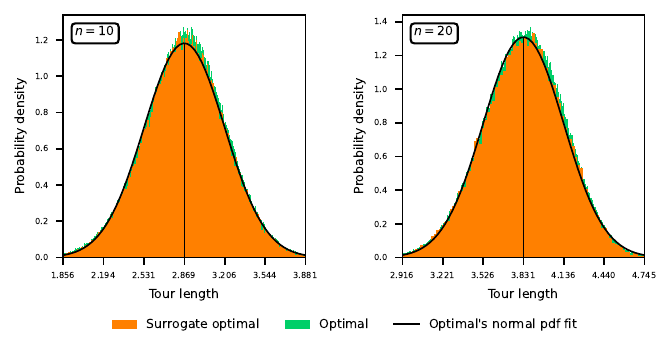}
    \caption{Histograms of surrogate tour lengths overlaying optimal tour lengths for 10-node and 20-node dataset pairs. Both are clearly normally distributed, justifying the t-test performed in Table~\ref{tab:proxyt}, and the two distributions are visually similar with the optimal histogram’s probability density function (pdf) fit aligning closely with the surrogate histogram. X-ticks are arranged in standard deviations of the optimality pdf fit.}
    \label{fig:proxynorm}
\end{figure}

\subsection{Dataset validation}
Near-optimal suboptimality scaling laws are sensitive to the precision of optimality estimates, so here we investigate approximation quality. For clarity, we refer to surrogate optimal solutions as “surrogate” and reserve “optimal” for strictly optimal solutions.

First, we compare surrogate and optimal solution distributions in 2D TSP where we have datasets for the latter via Concorde \cite{concordeweb}. Figure \ref{fig:proxynorm} shows that surrogate solution tour length closely approximates the distribution of optimal solution tour length at both 10 and 20 nodes using best-of-100 and best-of-1000 2-opt descents, respectively. Being normally distributed, we perform two sample t-tests on the means in Table \ref{tab:proxyt}. For each node scale, the mean surrogate tour length is slightly below the mean optimal tour length, which is possible since surrogate and optimal datasets sample distinct problems. Thus, along with the surrogate algorithm itself, we are testing whether the smaller size of the surrogate datasets can sufficiently approximate 1.28 million samples from the optimal distribution.

At both node scales, we are unable to reject the null hypothesis that mean surrogate tour length is equivalent to mean optimal tour length. Because these surrogate datasets underestimate optimality, from these results we clearly cannot suggest the surrogate mean is larger than the optimal mean. But we also do not find spurious evidence that the surrogate distribution underestimates the optimal distribution on average, with lower one-sided p-values of 0.377 and 0.183 at 10 and 20 nodes, respectively. With our chosen surrogate dataset settings, for 2D TSP with no more than 20 nodes, surrogate solutions appear statistically indistinguishable from optimal solutions (as desired).

\begin{table}[t]
    \caption{Two sample t-test comparing mean tour lengths of surrogate solutions and optimal solutions for 2D Euclidean TSP. P-values are obtained for both 10-node and 20-node dataset pairs. We fail to refute the null hypothesis that surrogate solutions come from the same distribution as optimal solutions.}
    \centering
    \begin{tabular}{rcccc}
        \toprule
        & \multicolumn{2}{c}{10 nodes} & \multicolumn{2}{c}{20 nodes} \\
        \cmidrule(r){2-3}
        \cmidrule(l){4-5}
        & Surrogate & Optimal & Surrogate & Optimal \\
        \midrule
        Samples & 128,000 & 1,280,000 & 64,000 & 1,280,000 \\
        Mean & 2.86839 & 2.86870 & 3.82970 & 3.83082 \\
        Standard deviation & 0.33727 & 0.33753 & 0.30534 & 0.30478 \\
        t-value & \multicolumn{2}{c}{-0.314} & \multicolumn{2}{c}{-0.906} \\
        \(\mathbb{P}(t<\text{t-value})\) & \multicolumn{2}{c}{0.377} & \multicolumn{2}{c}{0.183} \\
        \(\mathbb{P}(t>\text{t-value})\) & \multicolumn{2}{c}{0.623} & \multicolumn{2}{c}{0.817} \\
        \bottomrule
    \end{tabular}
    \label{tab:proxyt}
\end{table}

One problem would arise if the best-found local optimum (BLO) visitation frequency were overestimated by fewer descents in surrogate solution generation. This overestimation would imply surrogate solutions require more descents to find the true global optima located in relatively small basins of attraction. Instead, we find the BLO visitation rate of surrogate solution generation is almost identical to that observed in the small batch 100,000 descents-per-problem experiment in Figure \ref{fig:2opt_landscape} (center and top center). The error magnitudes do not exceed 1.5\% and 0.7\% for 10-node and 20-node dimension scaling arrays, respectively. Alongside the high BLO visitation rates themselves, this suggests global optima are often found with only a few 2-opt descents, a finding consistent with existing analysis of 3-opt \cite{tayarani2016analysis} (Figure 15 in cited work).

\begin{figure}[t]
    \centering
    \includegraphics[clip, trim=0in 0.1in 0in 0.08in, width=4.7in]{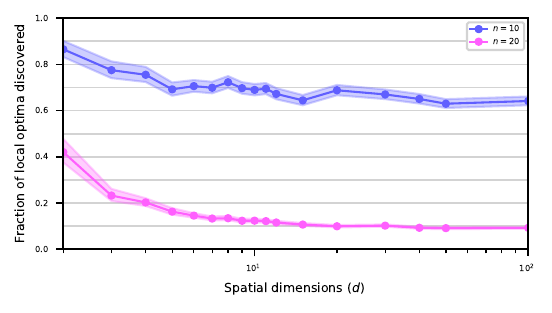}
    \caption{Surrogate solution generation’s estimated mean local optima discovery fraction. Fills show 99\% confidence intervals assuming local optima count is normally distributed. True mean local optima count is estimated with small batches of 2-opt local search performing 100,000 descents-per-problem. 10-node surrogate generation never falls below 60\% discovery, increasing confidence that best-found local optima are usually the true global optima. 20-node generation falls below 10\% discovery by 20 spatial dimensions, suggesting that best-found local optima at these scales may be suboptimal more often.}
    \label{fig:proxybigk}
\end{figure}

Another problem would arise if surrogate solution generation were to visit only a tiny fraction of local optima. Even with a high BLO visitation rate, eventually the number of missed global optima may accumulate. Figure~\ref{fig:proxybigk} plots the mean local optima discovery fraction of surrogate solution generation, where we estimate the true mean (divisor) using the 100,000 descents-per-problem experiments. The 10-node surrogate solutions uphold a greater than 60\% discovery rate by 100 dimensions, increasing confidence in global optima discovery considering the BLO visitation rate maintains values greater than 50\%. In contrast, 20-node surrogate solutions’ local optima discovery decreases to around 10\% by 100 dimensions. This discovery rate may still be sufficient to find the global optima in the overwhelming majority of problems. But given the 20-node BLO visitation rate sinks to roughly 6\% (Figure~\ref{fig:2opt_landscape} center), we are of course less confident. 

There are three smaller caveats for Figure \ref{fig:proxybigk} also worth noting. First, the confidence intervals assume that the number of local optima at each problem scale are normally distributed. We did not verify as we only saved summary statistics for these experiments. If the local optima count was (for instance) long-tailed toward larger values, this could add uncertainty to our estimates. The second caveat is that we ignore uncertainty in the surrogate solutions’ local optima discovery rate, which is relatively small since the surrogate datasets are orders of magnitude larger than those used to estimate the true discovery rate (2048 and 512 samples for 10 and 20 nodes, respectively). Lastly, 100,000 descents-per-problem may still miss elusive local optima with small basins of attraction. Even so, these local optima are quite unlikely to be global optima if 3-opt trends generalize to higher-dimensional 2-opt solutions \cite{tayarani2016analysis} (Figure 14 in cited work).

\FloatBarrier
\section{Compute requirements}
\label{apx:compute}
Each model was trained on an HPC node using one Nvidia Tesla V100 (16GB), 24 Intel Xeon Gold 6136 cores, and 384GB of memory. GPU usage was the bottleneck, so fewer CPU cores and less memory would be sufficient. As mentioned in Section~\ref{scalingmethod}, the 50-node RL run trained for 24 days and consumed roughly \mbox{\(3 \times 10^{-3}\)} PF-days of compute. The longest spatial dimension scaling runs and parameter scaling runs trained for about half that time and consumed around an order of magnitude less compute.

Optimal solution generation only used CPU compute and was performed on the same HPC nodes. Using Concorde \cite{concordeweb} via PyConcorde \cite{pyconcorde}, generating 1.28 million optimal solutions to 50-node 2D TSP required roughly 10 days. We provide 10 of these chunks for each TSP node scale we experimented on: 12.8 million solutions per scale and 128 million solutions in total. Generating approximately optimal datasets for higher-dimensional TSP (Section~\ref{optsolgen} and Appendix \ref{apx:proxyval}) required around 2 hours per 10-node scale and 60 hours per 20-node scale. Note that we used a Cython implementation of 2-opt, which we found to be several times faster than a pure Python implementation.

\FloatBarrier
\section{Model architecture}
\label{apx:arch}

Our model design prioritizes simplicity where possible. Besides making experiments easier to implement, this choice was intended to improve the reproducibility of our results and promote their generalization to other deep architectures. All core model components are implemented using PyTorch~\cite{paszke2017automatic} Transformer modules. Model code, algorithm code, and training scripts are provided in the project repository at \url{https://github.com/lowellw6/complexity-scaling-laws}. Specific hyperparameter choices are discussed separately in Appendix \ref{apx:hpo}.

\subsection{Policy network}
Our policy network is most similar to that introduced by Kool et al. \cite{kool2018attention} with two key simplifications. First, our encoder does not produce a full graph embedding, and so does not attend over a context node embedding during the subsequent forward passes of the autoregressive decoder. Second, unlike Kool et al. and Bresson et al. \cite{bresson2021transformernetworktravelingsalesman}, our decoder does not include a final single-head attention layer. Instead, our decoder performs multi-head attention over all node encodings for each decoding step while building a positionally-encoded partial tour in the decoder memory. No masking occurs until directly before logits are softmaxed and sampled, where probabilities are set to zero for previously selected nodes.

We moved forward with these simplifications after obtaining near-optimal performance without them at our experimented problem scales. But Kool et al.’s design has compelling motivation, especially concerning efficiency. Their graph embedding allows each decoder step to attend over the reduced subset of selectable nodes while still referencing a representation of the full problem. This linear ramp down of attention length still incurs \(O(n^3)\) complexity to decode a full tour, as does our approach which maintains constant attention length.\footnote{Self-attention has \(O(n^2)\) complexity w.r.t. to the node scale and attention length \(n\) \cite{vaswani2017attention}, which we iterate \(n\) times.} But pruning previously selected nodes requires only a third the compute, all else being equal. Given that autoregressive decoding is most of the forward pass, we suggest reconsidering this design choice when experimenting with large node scales.

\subsection{Critic network}
\begin{figure}
    \centering
    \includegraphics[width=5.5in]{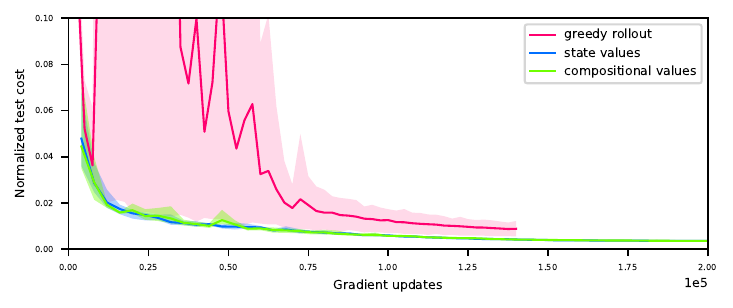}
    \caption{PPO baseline alternatives training on 2D 30-node TSP, evaluated on a held out set of 100,000 problems. We normalize suboptimality between optimal performance and random performance in the range \([0,1]\). We summarize 5 random seeds for each baseline, where lines showing the means and fills showing the min/max values. Performance converges well when learning problem-level (state) values or node-level (compositional) values with a critic network, with no meaningful difference between approaches. But using greedy rollouts via a separate policy (updated sparsely like a target Q-network) usually produces temporary divergence. We truncate this curve because two runs timed out due to external load on our HPC cluster. Of the three runs that reached 200,000 updates, none surpassed learned value function performance.}
    \label{fig:crt_abl}
\end{figure}

Our architecture learns values for the policy gradient baseline. Kool et al. instead use deterministic greedy rollouts from a baseline policy network, referencing the difficulty of multi-objective actor-critic learning. But we found the contrary to be true in our experiments. Our greedy rollout implementation often diverges early in training (Figure \ref{fig:crt_abl} pink), which may stem from one of several distinctions. First, we use a PPO actor loss rather than REINFORCE. Second, we synchronize the baseline policy every 625 gradient updates, matching Kool et al.’s period, but they only synchronize if a paired t-test determines statistically significant improvement. Lastly, we do not regenerate the evaluation dataset after synchronizing (to avoid overfitting), though we do use a larger samples size of 100,000. Where possible we also use hyperparameters from Appendix \ref{apx:hpo}, which are optimized for compositional value learning (discussed below), creating a less fair comparison.

Regardless, learning a value function was useful for studying critic loss trends (Appendix~\ref{apx:loss}). To learn problem-level state values, we use a second Transformer decoder which receives node encodings from the (now shared) policy encoder. But encodings are input as decoder memory (without positional encoding) and a zero-fill start token is input for the target.  The output is projected into a scalar value estimate which we train to predict on-policy tour cost for PPO, and optimal tour cost for SFT. Requiring just one forward pass, this is a relatively lightweight addition. We found this simple approach converges well in our experiments (Figure \ref{fig:crt_abl} blue).

However, problem-level value estimates provide an increasingly stale baseline the longer decoding iterates. For example, if just two nodes remain unvisited in a partial tour of 50, the range of possible cost outcomes dramatically reduces. From the perspective of the 49\textsuperscript{th} decoding, all previous node selections can be considered as components of the current node-level state. Note that our framework of problem-level states is merely the outcome of our \textit{choice} to formalize the TSP environment as a bandit problem. We could have instead chosen to formalize the environment as an MDP with node-level decisions, and if we set a discount factor \mbox{\(\gamma=1\)}, this MDP generates identical returns to our bandit problem framing. (Because no rewards occur before the terminal step.)

Attempting to learn more granular credit assignment, we implemented node-level value estimation. We refer to this alternative as \textit{compositional value learning}, and refer to problem-level value learning as state value learning. We compute compositional values through a straightforward modification to our state-value critic architecture: concatenating the positionally-encoded node selection sequence to the start token for target input. This produces \(n+1\) compositional values, each attending over incremental partial tours in one forward pass via a causal mask. The first output remains the problem-level state-value, attending only over the start token and node encodings. The final output is the Q-value, attending over all node selections. We explain how compositional values influence learning objectives in the following subsection.

\subsection{PPO using a compositional value baseline}
This section describes how we set up the PPO actor-critic loss to learn and leverage node-level compositional value estimates. We use this style of PPO for all RL experiments in the main paper. We omit these details for simplicity and because we observe no evidence of performance differences between node- and problem-level baselines (Figure \ref{fig:crt_abl} green and blue).

To leverage compositional values in the actor loss, the PPO clipped surrogate objective \cite{schulman2017proximalpolicyoptimizationalgorithms}, for each node selection we substitute the problem-level value for the node-specific value of the preceding partial tour. Formally, we minimize:

\begin{gather*}
    L^{CLIP}(\theta) = \E_{t,i} \left[ \max \left( r_{t,i}(\theta) A_{t,i}, \text{clip} (r_{t,i}(\theta), 1-\epsilon, 1+\epsilon) A_{t,i} \right) \right] \\
    A_{t,i} = c_t - V_{t,i}(\theta)
\end{gather*}

for on-policy tour \(t\) and node selection index \(i\), where we instead take the maximum term since we are minimizing cost \(c_t\). The ratio function \(r_{t,i}\) is unchanged besides specifying the policy's node selection index. Note that we do not add an entropy bonus since optimal edge selection is deterministic, and doing so may incentivize suboptimal infinite-compute performance, entangling the entropy objective with the model limitations we intend to observe. 

Modifying the critic loss to learn compositional values is less straightforward since cost is defined only for a complete solution. Transitioning from problem- to node-level credit assignment resembles the transformation from top MDP to bottom MDP illustrated by Metz et al.~\cite{metz2019discretesequentialpredictioncontinuous}.\footnote{Their related work section provides a useful introduction to other deep learning algorithms which address large discrete action spaces.} Because the return is shared between node selections within a particular tour, the corresponding compositional value targets are identical as well. Our critic loss is then:

\begin{gather*}
    L^{CRT}(\theta) = \E_{t,i} \left[ \left( c_t - V_{t,i}(\theta) \right)^2 \right]
\end{gather*}

In theory, this loss still allows compositional value targets to diversify in expectation. If rollouts from a given partial tour usually result in low-cost solutions, the expected compositional value target is small, and the inverse is also true. In preliminary experiments, compositional value learning appeared slightly more sample efficient compared to strictly learning problem-level state values, motivating our selection of the former. 

However, this distinction disappears when carefully controlling for other variables and averaging over several seeds, at least for 30-node TSP. After initial transient effects, performance distributions are indistinguishable (Figure \ref{fig:crt_abl}), and we noticed that compositional value estimates vary little within a given tour. These findings are not surprising given the critic attempts to learn granular values with a coarse reward signal, an analogue to value estimation in a sparsely rewarded MDP. Further,~given~tours are sampled on-policy for problems generated online, the critic sees each partial tour exactly once.

\FloatBarrier
\section{Hyperparameter settings}
\label{apx:hpo}
Here we provide the hyperparameter (HP) settings used for model training (Table \ref{tab:hps}), along with how we determined them through hyperparameter optimization (HPO). We optimize HP selection for PPO experiments, but we also use the obtained settings for SFT experiments where applicable.

Critically, previous work suggests scaling law fits are relatively insensitive to model and algorithm HPs like Transformer model shape and learning rate decay \cite{kaplan2020scalinglawsneurallanguage}. Within reasonable bounds, we do not expect different HP settings to severely alter the trends found in this paper so long as training sufficiently converges.

Setting HPs effectively by hand would be tedious and rely heavily on intuition. Reinforcement learning often involves more HPs than supervised learning, and setting model HPs for our custom architecture lacks the grounding of existing results. Algorithmic HPO has the added benefit of reducing wall-clock time that is required to reach sufficient convergence. With finite resources, unbottlenecked compute can only be approximated in the scaling laws that require it. Some of our intuition-informed preliminary experiments required several million gradient updates (and multiple learning rate warm restarts) to sufficiently converge, dramatically slowing research flow. Our HPO-informed experiments converged well in less than half the gradient updates.

\begin{table}[h]
    \caption{Hyperparameters used for RL node-scaling experiments. Other experiments, including SFT training runs, default to these settings where applicable unless otherwise specified. After learning rate cosine decay finishes, RL runs switch to a slow linear decay for the remainder of training.}
    \centering
    \begin{tabular}{lllc}
        \toprule
        Category & Name & Value & HPO-informed \\
        \midrule
        \multirow{4}{*}{PPO} & Minibatch size & 175 & No  \\
        & Minibatches & 4 & Yes  \\
        & Ratio clip & 0.17 & Yes  \\
        & Critic loss coefficient & 0.52 & Yes  \\
        \midrule
        \multirow{2}{*}{Optimizer} & Algorithm & Adam & No  \\
        & Gradient norm clip & 0.24 & Yes  \\
        \midrule
        \multirow{4}{*}{Scheduler} & Max learning rate & \(\text{9.37} \times \text{10}^\text{-5}\) & Yes \\
        &  Linear warm-up updates & 3,000 & No \\
        & Cosine decay finish update & 170,000 & Yes \\
        & Cosine decay floor & \(\text{10}^\text{-5}\) & No \\
        \midrule
        \multirow{7}{*}{Model} & Encoder layers & 3 & Yes \\
        & Policy decoder layers & 2 & Yes \\
        & Critic decoder layers & 2 & Yes \\
        & Transformer width & 184 & Yes \\
        & Transformer feedforward & 736 & No \\
        & Transformer attention heads & 8 & No \\
        & Transformer dropout & 0 & No \\
        \bottomrule
    \end{tabular}
    \label{tab:hps}
\end{table}

\subsection{HPO method}
We use BOHB \cite{bohb} implemented with Optuna \cite{optuna} to optimize update efficiency for 50-node TSP. We truly wish to optimize serial-compute efficiency, so we more strictly limit model depth, especially for the autoregressive policy decoder. We chose our maximum scale of 50 nodes to optimize the slowest run, and because we expect high-performing HPs for harder TSP scales to generalize well to smaller, easier scales. But HPs optimized for small scales may fail to converge on larger scales within the same update budget.

We chose BOHB for its balance of anytime and final performance \cite{bohb}. BOHB is no longer state-of-the-art in final performance but has comparable anytime performance \cite{smacv3} and is relatively easy to implement. We implemented BOHB using Optuna 3.5’s multivariate Tree-structured Parzen Estimator (TPE) \cite{NIPS2011_86e8f7ab} for HP sampling (the Bayesian optimization “BO” in BOHB) paired with their Hyberband \cite{JMLR:v18:16-558} pruning algorithm (the “HB” in BOHB). Table \ref{tab:bohb_setup} details the HPs we used for the BOHB algorithm itself along with our evaluation setup. We found BOHB especially sensitive to the minimum gradient updates budgeted per trial. Setting this value too low results in pruning based on noise before initial convergence. Setting too high wastes a significant amount of time and compute.

Several HPs in Table \ref{tab:hps} were fixed for various reasons. PPO minibatch size was maximized under GPU memory constraints for our search space. The learning rate linear warm-up schedule was fixed to avoid incentivizing rapid ascents that are more likely to diverge. Cosine learning rate decay was selected because we observed other schedules like exponential decaying too quickly to sufficiently converge; this issue was also noted in related work \cite{kaplan2020scalinglawsneurallanguage}. We fixed the learning rate decay floor to allow for a subsequent linear decay to zero over the remainder of training. This design choice couples fast, HPO-informed early convergence with meticulous late-stage convergence that appears to better approximate the infinite-compute performance limit.

Table \ref{tab:bohb_search} details our BOHB search space. We set relaxed bounds in preliminary attempts then trimmed unpromising regions. Runs often diverged when surpassing the reported upper bounds for max learning rate, gradient norm clip, and PPO ratio clip. GPU memory constraints limited model width's upper bound and our results suggest wider networks would perform better (Figure \ref{fig:hpo_bar} top-left).

Lastly, we found that forcing a warm-start, human-intuition HP configuration for trial 0 allows BOHB to find high-performing regions of the search space considerably faster than when using only random startup trials.

\subsection{BOHB results}
\begin{figure}
    \centering
    \includegraphics[clip, trim=0.5in 0in 0in 0in, width=1.00\textwidth]{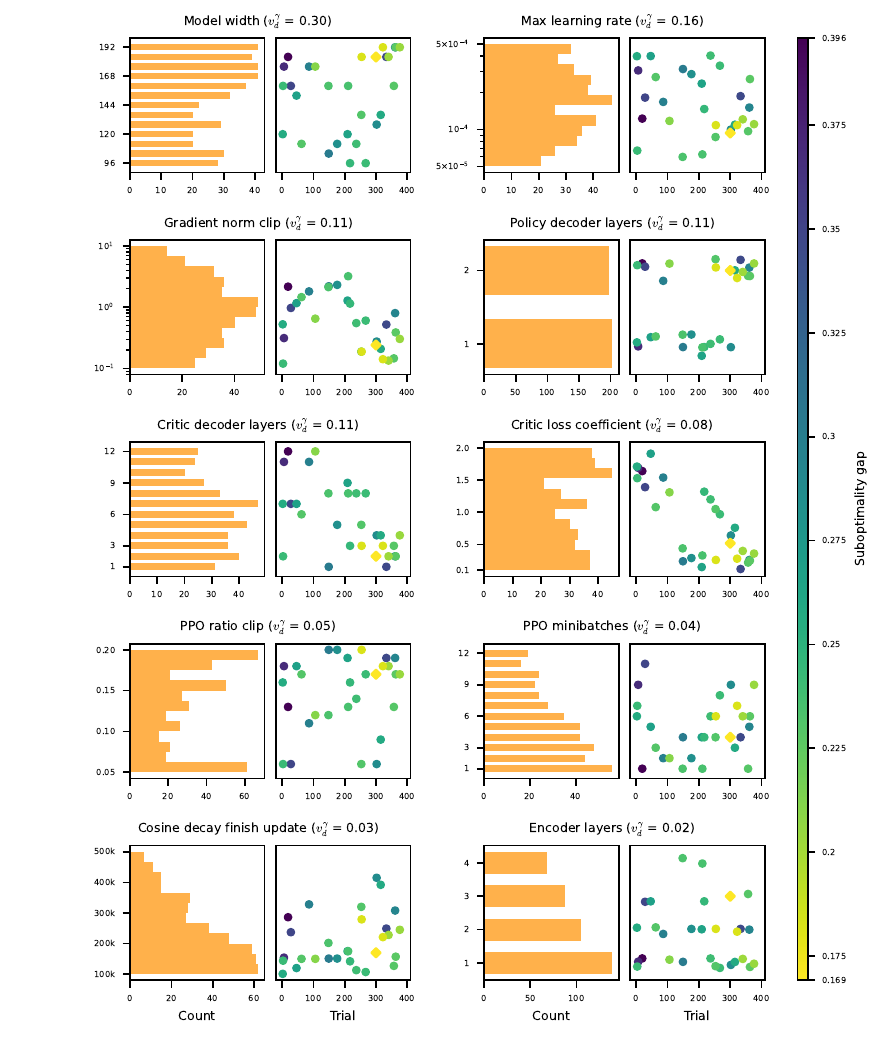}
    \caption{Search results of BOHB HPO on 50-node TSP for each HP in the search space. Y-axis labels are placed above subplots for clarity. The bar plots show HP search frequency over either discrete (bar gaps) or binned (no bar gaps) ranges, referencing all 400 trials. The scatter plots illustrate the performance of these HP ranges through the final scores of non-outlier completed trials, similarly to the plot generated by Optuna’s visualization.plot\_slice() function. Random Y-jitter is added to the policy decoder layers and encoder layers scatter plots to reduce point overlap. The best performing trial 300 is marked by a diamond (the wholeness of this number is just a coincidence and irrelevant to the HP sampling procedure). HPs are ranked by PED-ANOVA importance \(v_d^\gamma\) \cite{ijcai2023p0488} in descending order left to right, top to bottom. Note the distinctive clusters of high-performing (warmer) regions found over time for higher-importance HPs.}
    \label{fig:hpo_bar}
\end{figure}

Figure \ref{fig:hpo_bar} summarizes the BOHB search results that were used to inform our HP selection. Using larger model widths, a tight range near \(10^{-4}\) for max learning rate, and a small gradient norm clip less than 1 are most important for achieving high performance with fewer model updates. Convergence was particularly sensitive to gradient norm clip values greater than 1. In preliminary attempts without gradient clipping, pruning simply selected trials which most recently diverged.

PED-ANOVA importance ranking \cite{ijcai2023p0488} was chosen over the more common f-ANOVA \cite{pmlr-v32-hutter14} algorithm as we found the latter to be overly sensitive to poor performing regions of HP space (attributing majority importance to gradient norm clip). PED-ANOVA importance quantifies a HP’s degree of influence in attaining a top quantile of performance \cite{ijcai2023p0488}, which we found produces a more balanced ranking that also roughly aligns with concentration of high-performing clusters. We used a top-0.22 quantile which computed the local HP importance of the top 7 of 32 complete trials.

Optuna tracks objective scores by the final value achieved at the maximum update budget, not the best value achieved over the trial, which poses a question of statistical significance in near-optimal improvements. For example, trial 300 had the best outcome with a suboptimality gap of 0.169 at completion. But its minimum validation score over training was slightly below this at 0.166, though this happens to be the minimum score reported over all validations for all trials, which is encouraging. We evaluated the impact of our validation dataset size relative to these small performance deltas. This evaluation was performed retrospectively using the 50-node model from our main experiments since no checkpoints were saved during HPO. We rescored with 100 randomly sampled datasets of 10,000 problems each, obtaining a standard deviation of less than 0.003 suboptimality. The difference in suboptimality between the top two trials is over four standard deviations. Even comparing the final validation score of the top trial with the minimum score of the second-place trial returns a difference of over two standard deviations, demonstrating that our sample size is statistically meaningful assuming validation scores are normally distributed.

\newpage

\begin{table}
    \caption{BOHB algorithm and data settings. Remaining configurations use the defaults in Optuna 3.5.0. PPO minibatch size is fixed at 144 due to GPU memory constraints, and Transformer feedforward dimensions are set at \(4 \, d_{model}\). Otherwise, non-HPO-informed settings match Table \ref{tab:hps}.}
    \centering
    \begin{tabular}{lll}
        \toprule
        Category & Name & Value \\
        \midrule
        \multirow{4}{*}{Evaluation setup} & Trials & 400 \\
        & TSP nodes & 50 \\
        & Validation dataset size & 10,000  \\
        & Validation period (gradient updates) & 100  \\
        \midrule
        \multirow{3}{*}{Hyperband pruner} & Reduction factor \(\eta\) & 3 \\
        & Min gradient updates & 1,000 \\
        & Max gradient updates & 100,000 \\
        \midrule
        \multirow{2}{*}{TPE sampler} & Startup trials & 10 \\
        & Multivariate KDE & True \\
        \bottomrule
    \end{tabular}
    \label{tab:bohb_setup}
\end{table}

\begin{table}
    \caption{Experiment variables and corresponding search spaces for BOHB HPO. Continuous intervals are log-transformed. Transformer width's interval is the minimum value for 8 attention heads.}
    \centering
    \begin{tabular}{lllll}
        \toprule
        Category & Name & Min & Max & Interval \\
        \midrule
        \multirow{3}{*}{PPO} & Minibatches & 1 & 12 & 1 \\
        & Ratio clip & 0.05 & 0.2 & 0.01  \\
        & Critic loss coefficient & 0.1 & 2.0 & 0.01  \\
        \midrule
        Optimizer & Gradient norm clip & 0.1 & 10 & Continuous  \\
        \midrule
        \multirow{2}{*}{Scheduler} & Max learning rate & \(\text{5} \times \text{10}^\text{-5}\) & \(\text{5} \times \text{10}^\text{-4}\) & Continuous \\
        & Cosine decay finish update & 100,000 & 500,000 & 1,000  \\
        \midrule
        \multirow{4}{*}{Model} & Encoder layers & 1 & 4 & 1  \\
        & Policy decoder layers & 1 & 2 & 1  \\
        & Critic decoder layers & 1 & 12 & 1  \\
        & Transformer width & 96 & 192 & 8  \\
        \bottomrule
    \end{tabular}
    \label{tab:bohb_search}
\end{table}

\FloatBarrier

\FloatBarrier

\section{Proof of scaling relations and limits}
\label{apx:proofs}
Here we prove the node and spatial dimension scaling relations asserted in Section~\ref{pcs} mostly through extensions of existing analysis. We exclusively use the term “tour length” rather than “cost” for these demonstrations to maintain precise language.

For node scaling in 2D TSP, we simply show linear growth for both the expectation and variance of random tour length. The Beardwood–Halton–Hammersley Theorem \cite{beardwood1959shortest} already shows that expected optimal tour length grows proportionally to \(\sqrt{n}\) as \(n \to \infty\).

For spatial dimension scaling with a fixed number of nodes, we show that expected random tour length is strictly upper bounded by growth proportional to \(\sqrt{d}\). We then show that expected random tour length itself grows proportionally to \(\sqrt{d}\) as \(d \to \infty\) while random tour length variance and random tour suboptimality both approach a constant value.

\newtheorem{theorem}{Theorem}
\newtheorem{corollary}{Corollary}[theorem]
\newtheorem{lemma}[theorem]{Lemma}
\renewcommand\qedsymbol{$\blacksquare$}
\DeclarePairedDelimiter{\norm}{\lVert}{\rVert}
\newcommand\numberthis{\addtocounter{equation}{1}\tag{\theequation}}

\phantomsection
\addcontentsline{toc}{subsection}{Lemma 1}
\vspace{1\baselineskip}
\begin{lemma}
    Expected random tour length grows linearly w.r.t. \(n\).
    \label{proof:expn}
\end{lemma}
\begin{proof}
    Assume \(n\) node coordinates are sampled within a unit square such that each coordinate dimension is i.i.d. randomly sampled \(x_i \in [0,1]; \forall i \in \{1, 2\}\). Let \(\chi_i^{(p,q)} \in [-1, 1]\) be the random variable describing the difference between nodes \(p\) and \(q\) along the \(i^{th}\) axis:
    \begin{gather*}
        \chi_i^{(p,q)} = x_i^{(p)} - x_i^{(q)}
    \end{gather*}
    The Euclidean norm between nodes \(p\) and \(q\) is then
    \begin{gather*}
        \norm{X}_2^{(p,q)} = \sqrt{{\left( \chi_1^{(p,q)}\right)}^2 + {\left( \chi_2^{(p,q)} \right)}^2}
    \end{gather*}
    and we describe the full length of a random tour through each sampled point as
    \begin{gather*}
        \norm{X}_2^T = \sum_{k=1}^n  \norm{X}_2^{(p_k,q_k)} \quad \text{s.t. } p_k = q_{k-1},\; p_1 = q_n
    \end{gather*}
    Note that each \(\norm{X}_2^{(p_k,q_k)}\) variable shares sampled coordinates with their adjacent edges in the sum, meaning \(\norm{X}_2^T\) is \textit{not} a sum of independent variables. But these random edge norms are still identically distributed given that the choice of starting node \(p_1\) is arbitrary. Thus,
    \begin{gather*}
        \E \left[ \norm{X}_2^T \right] = \sum_{k=1}^n \E \left[ \norm{X}_2^{(p_k,q_k)} \right] = n \E \left[ \norm{X}_2^{(p,q)} \right] = n \, \mu_{\mathsmaller{\norm{X}}} \numberthis \label{eq:randn}
    \end{gather*}
    where \(\mu_{\mathsmaller{\norm{X}}}\) is the expected Euclidean distance between arbitrary nodes \(p\) and \(q\).
\end{proof}

For the uniform coordinate sampling used in this paper, evaluating \(\mu_{\mathsmaller{\norm{X}}}\) analytically involves a cumbersome quadruple integral for which several informal solutions can be found online. The exact solution is
\begin{gather*}
    \mu_{\mathsmaller{\norm{X}}} = \frac{2 + \sqrt{2} + 5 \ln(\sqrt{2} + 1)}{15} \approx 0.521
\end{gather*}
which our empirical fit approximates with trivial error.

\phantomsection
\addcontentsline{toc}{subsection}{Lemma 2}
\vspace{1\baselineskip}
\begin{lemma}
    Variance of random tour length grows linearly w.r.t. \(n\).
    \label{proof:varn}
\end{lemma}
\begin{proof}
    Because \(\norm{X}_2^T\) is \textit{not} a sum of independent random variables,
    \begin{gather*}
        \Var \left[ \norm{X}_2^T \right] \neq  \sum_{k=1}^n \Var \left[ \norm{X}_2^{(p_k,q_k)} \right]
    \end{gather*}
    This slightly complicates the proof. However, we can make use of the fact that dependencies are limited to adjacent edge variables. We start with the following equation:
    \begin{gather*}
        \Var \left[ \norm{X}_2^T \right] = \Var \left[ \sum_{k=1}^n \norm{X}_2^{(p_k,q_k)} \right] = \sum_{k=1}^n \sum_{j=1}^n \Cov \left( \norm{X}_2^{(p_k,q_k)}, \norm{X}_2^{(p_j,q_j)} \right)
    \end{gather*}
    which is true for any sum of random variables.\footnote{Derived from the equality \(\Var \left[ \sum_{i=1}^n X_i\right] = \E \left[ \left( \sum_{i=1}^n X_i \right)^2 \right] - \E \left[ \sum_{i=1}^n X_i \right]^2\)} Since \(\norm{X}_2^{(p_k,q_k)}\) only depends on \(\norm{X}_2^{(p_{k-1},q_{k-1})}\) and \(\norm{X}_2^{(p_{k+1},q_{k+1})}\) (with circular indexing for edges \(k \in \{1, n\}\)), all non-adjacent edge combinations have zero covariance. Thus, the previous expression simplifies to
    \begin{align*}
        \sum_{k=1}^n &\left[ \Cov \left(\norm{X}_2^{(p_k,q_k)}, \norm{X}_2^{(p_{k-1},q_{k-1})} \right) \right. \\
        &+ \Cov \left(\norm{X}_2^{(p_k,q_k)}, \norm{X}_2^{(p_k,q_k)}\right) \\
        &+ \Cov \left. \left(\norm{X}_2^{(p_k,q_k)}, \norm{X}_2^{(p_{k+1},q_{k+1})}\right) \right]
    \end{align*}
    Given that covariances between adjacent edge lengths are identically distributed, we obtain
    \begin{align*}
        \Var \left[ \norm{X}_2^T \right] &= \sum_{k=1}^n \left[ \Var \left(\norm{X}_2^{(p_k,q_k)} \right) + 2 \Cov \left(\norm{X}_2^{(p_k,q_k)}, \norm{X}_2^{(p_{k+1},q_{k+1})} \right) \right] \\
        &= n \left[ \Var \left(\norm{X}_2^{(p,q)}\right)  + 2 \Cov \left( \norm{X}_2^{(p,q)}, \norm{X}_2^{(q,s)}  \right) \right] \numberthis \label{eq:randvarn} \\
        &= n \left( \xi^{(p,q,s)} \right)^2
    \end{align*}
    where \((p,q,s)\) denotes an arbitrary 3-node sequence in the tour with edges \((p,q)\) and \((q,s)\), and \mbox{\(\left( \xi^{(p,q,s)} \right)^2\) is defined as the summation term obtained above}.
\end{proof}

Hence, the standard deviation of random tour length grows proportionally to \(\sqrt{n}\), which our empirical results closely approximate (Figure \ref{fig:rawstd} top left, \(\alpha \approx 0.497\)).

We could not find existing solutions for \(\xi^{(p,q,s)}\) and make no attempt to evaluate it ourselves. But our empirical fit suggests \(\xi^{(p,q,s)} \approx 0.276\) for uniform coordinate sampling.

\phantomsection
\addcontentsline{toc}{subsection}{Lemma 3}
\vspace{1\baselineskip}
\begin{lemma}
    Expected random tour length w.r.t. \(d\) is upper bounded by a function proportional to \(\sqrt{d}\).
    \label{proof:boundd}
\end{lemma}
\begin{proof}
    In a unit hypercube of \(d\) spatial dimensions consider a fixed number of \(n\) nodes i.i.d. randomly sampled as previously described. First, we obtain the expected value of the square of the high-dimensional Euclidean norm between two random nodes:
    \begin{gather*}
        \E \left[ \left( \norm{X}_2^{(p,q)} \right)^2 \right] = \E \left[ \sum_{k=1}^d \left( \chi_k^{(p,q)} \right)^2 \right] = d \, \E \left[ \left( \chi^{(p,q)} \right)^2 \right] = d \, \mu_{\mathsmaller{\chi^2}}
    \end{gather*}
    where \(\mu_{\mathsmaller{\chi^2}} \in [0,1]\) is the expected squared difference between \(p\) and \(q\) along an arbitrary spatial dimension axis. From the concave form of Jensen’s Inequality, it follows that
    \begin{gather*}
        \E \left[ \norm{X}_2^{(p,q)} \right] = \E \left[ \sqrt{\left( \norm{X}_2^{(p,q)} \right)^2} \right] \leq \sqrt{\E \left[ \left( \norm{X}_2^{(p,q)} \right)^2 \right] } = \sqrt{d \, \mu_{\mathsmaller{\chi^2}}}
    \end{gather*}
    Thus,
    \begin{gather*}
        \E \left[ \norm{X}_2^{(p,q)} \right] \leq \sqrt{d \, \mu_{\mathsmaller{\chi^2}}}
    \end{gather*}
    Note that \(\sqrt{d}\) is the hypercube’s maximal Euclidean norm, measuring from one corner to its farthest opposing corner. Hence the expected Euclidean distance between two randomly sampled nodes is bounded below that maximal length scaled by \(\sqrt{\mu_{\mathsmaller{\chi^2}}}\). Since Equation \ref{eq:randn} generalizes to higher dimensions, extending this result to full tour length is simple:
    \begin{gather*}
        \E \left[ \norm{X}_2^T \right] = n \, \E \left[ \norm{X}_2^{(p,q)} \right] \leq n \sqrt{d \, \mu_{\mathsmaller{\chi^2}}}
    \end{gather*}
    So,
    \begin{gather*}
        \E \left[ \norm{X}_2^T \right] \leq n \sqrt{\mu_{\mathsmaller{\chi^2}} \, d}
    \end{gather*}
\end{proof}

\begin{figure}[h]
    \centering
    \includegraphics[clip, trim=0in 0.1in 0in 0in, width=4.0in]{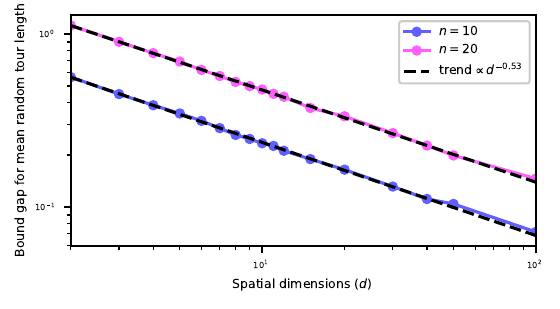}
    \caption{Upper bound for expected random tour length subtracted by observed means for both the 10-node and 20-node spatial dimension scaling experiments. Means evaluate 128,000 and 64,000 samples, respectively. This gap below the upper bound asymptotically decays toward zero (provably). Between node scales, decay rates are both close to square root power (\(\alpha \approx 0.53\)) so increasing the number of nodes simply increases the proportionality constant from roughly 0.82 to 1.62. This is notably close to the proportionality of node increase (doubling).}
    \label{fig:bound_approach}
\end{figure}

\begin{figure}[h]
    \centering
    \includegraphics[clip, trim=0.05in 0.1in 0.1in 0.1in, width=3.0in]{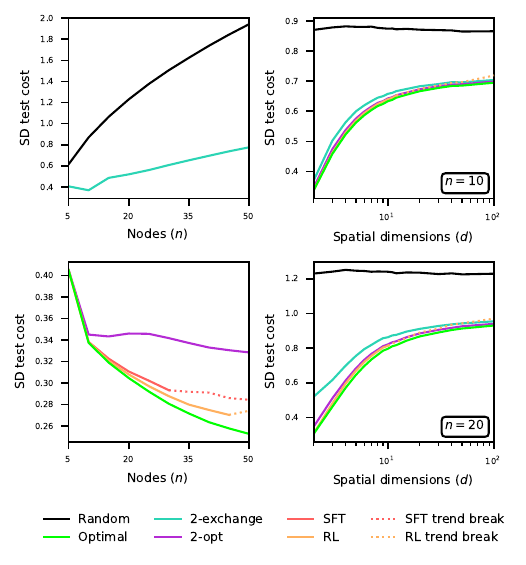}
    \caption{Standard deviation (SD) of cost over TSP node and spatial dimension scaling, shown for random, optimal, and algorithm distributions. Behavior of random tour length variance aligns with our analysis in Lemma~\ref{proof:varn} and Theorem~\ref{proof:limvard}. \textbf{Top left:} Tour length SD for random performance and mediocre algorithms like 2-exchange diverge w.r.t. number of nodes. \textbf{Bottom left:} Tour length SD for sufficiently near-optimal distributions decreases w.r.t. nodes, implying optimal tours become more similar in length with increasing node density in the representation space (at least for 2D TSP). \textbf{Right:} Like random tour suboptimality, random tour variance w.r.t. spatial dimensions is roughly constant over the tested domain and provably constant in the limit (Theorem~\ref{proof:limvard}). All other distributions produce an increasing convergent trend.}
    \label{fig:rawstd}
\end{figure}

If we wish to specify this bound for the uniform coordinate sampling used in this paper, it is straightforward to obtain \(\mu_{\mathsmaller{\chi^2}}\) directly from the continuous definition of expectation. Let \(f\) be a random variable’s probability density function and \(\rho\) and \(\omega\) the integrand variables for \(x_i^{(p)}\) and \(x_i^{(q)}\), respectively. Then,
\begin{gather*}
    \mu_{\mathsmaller{\chi^2}} = \E \left[ \left( \chi^{(p,q)} \right)^2 \right] = \int_0^1 \int_0^1 f(\omega) f(\rho) (\rho - \omega)^2 \, d\rho \, d\omega
\end{gather*}
Since \(f(\rho)\) and \(f(\omega)\) are uniform between 0 and 1, this simplifies to
\begin{gather*}
    \mu_{\mathsmaller{\chi^2}} = \int_0^1 \int_0^1 (\rho - \omega)^2 \, d\rho \, d\omega = \int_0^1 \left(\omega^2 - \omega + \frac{1}{3} \right) d\omega = \frac{1}{6}
\end{gather*}
Thus, for any TSP problem distribution in this paper,
\begin{gather*}
    \E \left[ \norm{X}_2^T \right] \leq n \sqrt{\frac{d}{6}}
\end{gather*}
We plot the difference between this upper bound and our empirical dimension-scaling results in Figure \ref{fig:bound_approach}. Interestingly, this gap also forms a smooth power law.

\phantomsection
\addcontentsline{toc}{subsection}{Theorem 4}
\vspace{1\baselineskip}
\begin{theorem}
    Expected random tour length is proportional to \(\sqrt{d}\) as \(d \to \infty\).
    \label{proof:limd}
\end{theorem}
\begin{proof}
    Extending existing analysis, it is simple to show that the previous upper bound becomes arbitrarily tight in the limit as \(d \to \infty\). Specifying the Euclidean norm case\footnote{\(p=2\) in the cited work. Our usage of \(p\) as the norm starting node is unrelated.} for the proof of Lemma 1 in François et al. \cite{franccois2007concentration} (Section 5.1.1), it directly follows that
    \begin{gather*}
        \lim_{d \to \infty} \frac{\E \left[ \norm{X}_2^{(p,q)} \right]}{\sqrt{d}} = \sqrt{\mu_{\mathsmaller{\chi^2}}} \numberthis \label{eq:limdexp}
    \end{gather*}
    Borrowing Equation \ref{eq:randn} once again, for the full tour length we obtain
    \begin{gather*}
        \lim_{d \to \infty} \frac{\E \left[ \norm{X}_2^T \right]}{\sqrt{d}} = \lim_{d \to \infty} \frac{n \E \left[ \norm{X}_2^{(p,q)} \right]}{\sqrt{d}} = n \lim_{d \to \infty} \frac{\E \left[ \norm{X}_2^{(p,q)} \right]}{\sqrt{d}} = n \sqrt{\mu_{\mathsmaller{\chi^2}}}
    \end{gather*}
    or
    \begin{gather*}
        \lim_{d \to \infty} \frac{\E \left[ \norm{X}_2^T \right]}{\sqrt{d}} = n \sqrt{\mu_{\mathsmaller{\chi^2}}}
    \end{gather*}
\end{proof}

For this paper’s uniform coordinate sampling distribution, this becomes
\begin{gather*}
    \lim_{d \to \infty} \frac{\E \left[ \norm{X}_2^T \right]}{\sqrt{d}} = \frac{n}{\sqrt{6}}
\end{gather*}
implying that the upper bound gaps observed in Figure \ref{fig:bound_approach} indeed converge toward zero.

\phantomsection
\addcontentsline{toc}{subsection}{Theorem 5}
\vspace{1\baselineskip}
\begin{theorem}
    Variance of random tour length approaches a constant value as \(d \to \infty\).
    \label{proof:limvard}
\end{theorem}
\begin{proof}
    For this limit we extend Lemma 2 in François et al. (proved in Section 5.1.2 of their paper). For the Euclidean norm case, their Lemma shows
    \begin{gather*}
        \lim_{d\to\infty} \Var \left[ \norm{X}_2^{(p,q)} \right] = \frac{\left( \sigma_{\mathsmaller{\chi^2}}\right)^2}{4 \, \mu_{\mathsmaller{\chi^2}}}
    \end{gather*}
    where \(\sigma_{\mathsmaller{\chi^2}}\) is the standard deviation\footnote{François et al. at first describe \(\sigma_{\mathsmaller{\chi^2}}\) as variance, but later in their proof show it to be standard deviation.} of \(\left( \chi^{(p,q)} \right)^2\) for an arbitrary spatial dimension. Since Equation \ref{eq:randvarn} holds for higher dimensions,
    \begin{align*}
        \Var \left[ \norm{X}_2^T \right] &= n \left[ \Var \left(\norm{X}_2^{(p,q)}\right)  + 2 \Cov \left( \norm{X}_2^{(p,q)}, \norm{X}_2^{(q,s)}  \right) \right] \\
        \lim_{d\to\infty} \Var \left[ \norm{X}_2^T \right] &= n \left[ \lim_{d\to\infty} \Var \left(\norm{X}_2^{(p,q)}\right)  + 2 \lim_{d\to\infty} \Cov \left( \norm{X}_2^{(p,q)}, \norm{X}_2^{(q,s)}  \right) \right] \\
        &= n \left[ \frac{\left( \sigma_{\mathsmaller{\chi^2}}\right)^2}{4 \, \mu_{\mathsmaller{\chi^2}}}  + 2 \lim_{d\to\infty} \Cov \left( \norm{X}_2^{(p,q)}, \norm{X}_2^{(q,s)}  \right) \right] \numberthis \label{eq:constvar1}
    \end{align*}
    We now follow reasoning analogous to François et al.'s initial steps proving Lemma 2 of their work, but for covariance instead of variance. By definition,
    \begin{gather*}
        \Cov \left( \norm{X}_2^{(p,q)}, \norm{X}_2^{(q,s)}  \right) = \E \left[ \left( \norm{X}_2^{(p,q)} - \E \left[ \norm{X}_2^{(p,q)} \right] \right) \left( \norm{X}_2^{(q,s)} - \E \left[ \norm{X}_2^{(q,s)} \right] \right) \right]
    \end{gather*}
    Thus, using the Fatou–Lebesgue theorem to move the limit inside expectation,
    \begin{align*}
        &\lim_{d\to\infty} \Cov \left( \norm{X}_2^{(p,q)}, \norm{X}_2^{(q,s)}  \right) \\ &= \E \left[ \lim_{d\to\infty} \left( \left( \norm{X}_2^{(p,q)} - \E \left[ \norm{X}_2^{(p,q)} \right] \right) \left( \norm{X}_2^{(q,s)} - \E \left[ \norm{X}_2^{(q,s)} \right] \right) \right) \right] \numberthis \label{eq:constvar2}
    \end{align*}
    We may rewrite the first product term as
    \begin{gather*}
        \norm{X}_2^{(p,q)} - \E \left[ \norm{X}_2^{(p,q)} \right] = \frac{\left( \norm{X}_2^{(p,q)} \right)^2 - \E \left[ \norm{X}_2^{(p,q)} \right]^2 }{\norm{X}_2^{(p,q)} - \E \left[ \norm{X}_2^{(p,q)} \right]} = \frac{\frac{\left( \norm{X}_2^{(p,q)} \right)^2 - \E \left[ \norm{X}_2^{(p,q)} \right]^2}{\sqrt{d}}}{\frac{\norm{X}_2^{(p,q)}}{\sqrt{d}} + \frac{\E \left[ \norm{X}_2^{(p,q)} \right]}{\sqrt{d}}} \numberthis \label{eq:constvar3}
    \end{gather*}
    Referencing Equation \ref{eq:limdexp}, along with François et al.'s intermediate step 1 result from their Lemma 1 proof, we know the limit of the denominator almost surely approaches a constant:
    \begin{gather*}
        \lim_{d\to\infty} \left( \frac{\norm{X}_2^{(p,q)}}{\sqrt{d}} + \frac{\E \left[ \norm{X}_2^{(p,q)} \right]}{\sqrt{d}} \right) = \lim_{d\to\infty} \frac{\norm{X}_2^{(p,q)}}{\sqrt{d}} + \lim_{d\to\infty} \frac{\E \left[ \norm{X}_2^{(p,q)} \right]}{\sqrt{d}}
    \end{gather*}
    and
    \begin{gather*}
        \mathbb{P} \left[ \lim_{d\to\infty} \frac{\norm{X}_2^{(p,q)}}{\sqrt{d}} = \sqrt{\mu_{\mathsmaller{\chi^2}}} \right] = 1
    \end{gather*}
    so
    \begin{gather*}
        \mathbb{P} \left[ \lim_{d\to\infty} \left( \frac{\norm{X}_2^{(p,q)}}{\sqrt{d}} + \frac{\E \left[ \norm{X}_2^{(p,q)} \right]}{\sqrt{d}} \right) = \sqrt{\mu_{\mathsmaller{\chi^2}}} + \sqrt{\mu_{\mathsmaller{\chi^2}}} = 2 \sqrt{\mu_{\mathsmaller{\chi^2}}} \right] = 1
    \end{gather*}
    Meanwhile, the limit of the numerator of Equation \ref{eq:constvar3} is simplified using Equation \ref{eq:limdexp} again. First, note that
    \begin{align*}
        \left( \norm{X}_2^{(p,q)} \right)^2 - \E \left[ \norm{X}_2^{(p,q)} \right]^2 &= \sum_{k=1}^d \left( \chi_k^{(p,q)}\right)^2 - \E \left[ \norm{X}_2^{(p,q)} \right]^2 \\ &= \mathlarger{\sum}_{k=1}^d \left( \left( \chi_k^{(p,q)}\right)^2 - \frac{\E \left[ \norm{X}_2^{(p,q)} \right]^2}{d} \right)
    \end{align*}
    and since
    \begin{gather*}
        \lim_{d\to\infty} \frac{\E \left[ \norm{X}_2^{(p,q)} \right]^2}{d} = \lim_{d\to\infty} \left( \frac{\E \left[ \norm{X}_2^{(p,q)} \right]}{\sqrt{d}} \right)^2 = \left( \lim_{d\to\infty} \frac{\E \left[ \norm{X}_2^{(p,q)} \right]}{\sqrt{d}} \right)^2 = \mu_{\mathsmaller{\chi^2}}
    \end{gather*}
    taking the limit and substituting yields
    \begin{align*}
        \lim_{d\to\infty} \left( \left( \norm{X}_2^{(p,q)} \right)^2 - \E \left[ \norm{X}_2^{(p,q)} \right]^2 \right) &= \lim_{d\to\infty} \mathlarger{\sum}_{k=1}^d \left( \left( \chi_k^{(p,q)}\right)^2 - \frac{\E \left[ \norm{X}_2^{(p,q)} \right]^2}{d} \right) \\ &= \lim_{d\to\infty} \sum_{k=1}^d \left( \left( \chi_k^{(p,q)}\right)^2 - \mu_{\mathsmaller{\chi^2}} \right)
    \end{align*}
    Returning to Equation \ref{eq:constvar2}, notice that all analysis from Equation \ref{eq:constvar3} up until this point produces analogous results for the second product term \(\left( \norm{X}_2^{(q,s)} - \E \left[ \norm{X}_2^{(q,s)} \right] \right)\) since \(\norm{X}_2^{(p,q)}\) and \(\norm{X}_2^{(q,s)}\) are identically distributed. Thus,
    \begin{align*}
        &\lim_{d\to\infty} \left( \left( \norm{X}_2^{(p,q)} - \E \left[ \norm{X}_2^{(p,q)} \right] \right) \left( \norm{X}_2^{(q,s)} - \E \left[ \norm{X}_2^{(q,s)} \right] \right) \right) \\ &= \lim_{d\to\infty} \left( \frac{\frac{\left( \norm{X}_2^{(p,q)} \right)^2 - \E \left[ \norm{X}_2^{(p,q)} \right]^2}{\sqrt{d}}}{\frac{\norm{X}_2^{(p,q)}}{\sqrt{d}} + \frac{\E \left[ \norm{X}_2^{(p,q)} \right]}{\sqrt{d}}} \frac{\frac{\left( \norm{X}_2^{(q,s)} \right)^2 - \E \left[ \norm{X}_2^{(q,s)} \right]^2}{\sqrt{d}}}{\frac{\norm{X}_2^{(q,s)}}{\sqrt{d}} + \frac{\E \left[ \norm{X}_2^{(q,s)} \right]}{\sqrt{d}}} \right) \\ &= \lim_{d\to\infty} \left( \frac{\frac{\left(\left( \norm{X}_2^{(p,q)} \right)^2 - \E \left[ \norm{X}_2^{(p,q)} \right]^2\right)\left(\left( \norm{X}_2^{(q,s)} \right)^2 - \E \left[ \norm{X}_2^{(q,s)} \right]^2\right)}{d}}{\left(\frac{\norm{X}_2^{(p,q)}}{\sqrt{d}} + \frac{\E \left[ \norm{X}_2^{(p,q)} \right]}{\sqrt{d}}\right) \left(\frac{\norm{X}_2^{(q,s)}}{\sqrt{d}} + \frac{\E \left[ \norm{X}_2^{(q,s)} \right]}{\sqrt{d}}\right)} \right)
    \end{align*}
    and with probability 1,
    \begin{align*}
        &= \frac{\lim\limits_{d\to\infty} \frac{\left(\left( \norm{X}_2^{(p,q)} \right)^2 - \E \left[ \norm{X}_2^{(p,q)} \right]^2\right)\left(\left( \norm{X}_2^{(q,s)} \right)^2 - \E \left[ \norm{X}_2^{(q,s)} \right]^2\right)}{d}}{\lim\limits_{d\to\infty} \left( \frac{\norm{X}_2^{(p,q)}}{\sqrt{d}} + \frac{\E \left[ \norm{X}_2^{(p,q)} \right]}{\sqrt{d}}\right) \lim\limits_{d\to\infty} \left( \frac{\norm{X}_2^{(q,s)}}{\sqrt{d}} + \frac{\E \left[ \norm{X}_2^{(q,s)} \right]}{\sqrt{d}}\right)} \\ &= \frac{\lim\limits_{d\to\infty} \frac{\left(\sum_{k=1}^d \left( \left( \chi_k^{(p,q)}\right)^2 - \mu_{\mathsmaller{\chi^2}} \right) \right)\left( \sum_{j=1}^d \left( \left( \chi_j^{(q,s)}\right)^2 - \mu_{\mathsmaller{\chi^2}} \right)\right)}{d}}{4 \, \mu_{\mathsmaller{\chi^2}}} \numberthis \label{eq:constvar4}
    \end{align*}
    Here, the numerator differs slightly from that in the proof from François et al. given we are proving for covariance rather than variance. But we can still apply reasoning analogous to their remaining steps. By the definition of covariance, and since \(\E \left[ \sum_{k=1}^d \left( \left( \chi_k^{(p,q)}\right)^2 - \mu_{\mathsmaller{\chi^2}} \right) \right] = 0\),
    \begin{align*}
        &\Cov \left( \sum_{k=1}^d \left( \left( \chi_k^{(p,q)}\right)^2 - \mu_{\mathsmaller{\chi^2}} \right), \sum_{j=1}^d \left( \left( \chi_j^{(q,s)}\right)^2 - \mu_{\mathsmaller{\chi^2}} \right) \right) \\ &= \E \left[ \left( \sum_{k=1}^d \left( \left( \chi_k^{(p,q)}\right)^2 - \mu_{\mathsmaller{\chi^2}} \right)\right) \left( \sum_{j=1}^d \left( \left( \chi_j^{(q,s)}\right)^2 - \mu_{\mathsmaller{\chi^2}} \right)\right) \right] \\ &\qquad - \E \left[\sum_{k=1}^d \left( \left( \chi_k^{(p,q)}\right)^2 - \mu_{\mathsmaller{\chi^2}} \right) \right] \E \left[\sum_{j=1}^d \left( \left( \chi_j^{(q,s)}\right)^2 - \mu_{\mathsmaller{\chi^2}} \right) \right] \\ &= \E \left[ \left( \sum_{k=1}^d \left( \left( \chi_k^{(p,q)}\right)^2 - \mu_{\mathsmaller{\chi^2}} \right)\right) \left( \sum_{j=1}^d \left( \left( \chi_j^{(q,s)}\right)^2 - \mu_{\mathsmaller{\chi^2}} \right)\right) \right] - 0
    \end{align*}
    Thus,
    \begin{align*}
        &\E \left[ \left( \sum_{k=1}^d \left( \left( \chi_k^{(p,q)}\right)^2 - \mu_{\mathsmaller{\chi^2}} \right)\right) \left( \sum_{j=1}^d \left( \left( \chi_j^{(q,s)}\right)^2 - \mu_{\mathsmaller{\chi^2}} \right)\right) \right] \\ &= \Cov \left( \sum_{k=1}^d \left( \left( \chi_k^{(p,q)}\right)^2 - \mu_{\mathsmaller{\chi^2}} \right), \sum_{j=1}^d \left( \left( \chi_j^{(q,s)}\right)^2 - \mu_{\mathsmaller{\chi^2}} \right) \right) \\ &= \sum_{k=1}^d \sum_{j=1}^d \Cov \left( \left( \left( \chi_k^{(p,q)}\right)^2 - \mu_{\mathsmaller{\chi^2}} \right), \left( \left( \chi_j^{(q,s)}\right)^2 - \mu_{\mathsmaller{\chi^2}} \right) \right)
    \end{align*}
    and since coordinates are independently sampled between dimensions,
    \begin{align*}
        &= \sum_{k=1}^d \Cov \left( \left( \left( \chi_k^{(p,q)}\right)^2 - \mu_{\mathsmaller{\chi^2}} \right), \left( \left( \chi_k^{(q,s)}\right)^2 - \mu_{\mathsmaller{\chi^2}} \right) \right) \qquad
    \end{align*}
    With the observation \(\E \left[ \left( \chi_k^{(p,q)}\right)^2 - \mu_{\mathsmaller{\chi^2}} \right] = 0\) alongside another recursion through the definition of covariance, we obtain
    \begin{align*}
        &\sum_{k=1}^d \Cov \left( \left( \left( \chi_k^{(p,q)}\right)^2 - \mu_{\mathsmaller{\chi^2}} \right), \left( \left( \chi_k^{(q,s)}\right)^2 - \mu_{\mathsmaller{\chi^2}} \right) \right) \\ &= \sum_{k=1}^d \left( \E \left[\left( \left( \chi_k^{(p,q)}\right)^2 - \mu_{\mathsmaller{\chi^2}} \right) \left( \left( \chi_k^{(q,s)}\right)^2 - \mu_{\mathsmaller{\chi^2}} \right) \right] \right. \\ &\qquad - \left. \E \left[\left( \chi_k^{(p,q)}\right)^2 - \mu_{\mathsmaller{\chi^2}} \right]   \E \left[\left(\chi_k^{(q,s)}\right)^2 - \mu_{\mathsmaller{\chi^2}} \right]\right) \\ &= \sum_{k=1}^d \left( \E \left[\left( \left( \chi_k^{(p,q)}\right)^2 - \mu_{\mathsmaller{\chi^2}} \right) \left( \left( \chi_k^{(q,s)}\right)^2 - \mu_{\mathsmaller{\chi^2}} \right) \right] - 0 \right) \\ &= d \left( \varphi^{(p,q,s)} \right)^2
    \end{align*}
    since the product inside the expected value is identically distributed for each summation term over \(d\). We define this summation term as \mbox{\(\left( \varphi^{(p,q,s)} \right)^2\)}. Condensing, we arrive at
    \begin{gather*}
        \E \left[ \left( \sum_{k=1}^d \left( \left( \chi_k^{(p,q)}\right)^2 - \mu_{\mathsmaller{\chi^2}} \right)\right) \left( \sum_{j=1}^d \left( \left( \chi_j^{(q,s)}\right)^2 - \mu_{\mathsmaller{\chi^2}} \right)\right) \right] = d \left( \varphi^{(p,q,s)} \right)^2 \numberthis \label{eq:constvar5}
    \end{gather*}
    Returning to Equation \ref{eq:constvar2}, we can now see that with probability 1,
    \begin{align*}
        &\lim_{d\to\infty} \Cov \left( \norm{X}_2^{(p,q)}, \norm{X}_2^{(q,s)}  \right) \\ &= \E \left[ \lim_{d\to\infty} \left( \left( \norm{X}_2^{(p,q)} - \E \left[ \norm{X}_2^{(p,q)} \right] \right) \left( \norm{X}_2^{(q,s)} - \E \left[ \norm{X}_2^{(q,s)} \right] \right) \right) \right] \\ &= \E \left[ \frac{\lim\limits_{d\to\infty} \frac{\left(\sum_{k=1}^d \left( \left( \chi_k^{(p,q)}\right)^2 - \mu_{\mathsmaller{\chi^2}} \right) \right)\left( \sum_{j=1}^d \left( \left( \chi_j^{(q,s)}\right)^2 - \mu_{\mathsmaller{\chi^2}} \right)\right)}{d}}{4 \, \mu_{\mathsmaller{\chi^2}}} \right] \quad \text{via Eq. \ref{eq:constvar4}} \\ &= \frac{\lim\limits_{d\to\infty} \frac{\E \left[ \left(\sum_{k=1}^d \left( \left( \chi_k^{(p,q)}\right)^2 - \mu_{\mathsmaller{\chi^2}} \right) \right)\left( \sum_{j=1}^d \left( \left( \chi_j^{(q,s)}\right)^2 - \mu_{\mathsmaller{\chi^2}} \right)\right)\right]}{d}}{4 \, \mu_{\mathsmaller{\chi^2}}} \quad \text{via Fatou–Lebesgue theorem} \\ &= \frac{\lim\limits_{d\to\infty} \frac{d \left( \varphi^{(p,q,s)} \right)^2}{d}}{4 \, \mu_{\mathsmaller{\chi^2}}} \quad \text{via Eq. \ref{eq:constvar5}} \\ &= \frac{\left( \varphi^{(p,q,s)} \right)^2}{4 \, \mu_{\mathsmaller{\chi^2}}}
    \end{align*}
    Thus,
    \begin{gather*}
        \mathbb{P} \left[ \lim_{d\to\infty} \Cov \left( \norm{X}_2^{(p,q)}, \norm{X}_2^{(q,s)} \right) = \frac{\left( \varphi^{(p,q,s)} \right)^2}{4 \, \mu_{\mathsmaller{\chi^2}}} \right] = 1
    \end{gather*}
    Referencing step 2 in François et al.'s proof of their Lemma 1, the same reasoning shows
    \begin{gather*}
        \lim_{d\to\infty} \Cov \left( \norm{X}_2^{(p,q)}, \norm{X}_2^{(q,s)} \right) = \frac{\left( \varphi^{(p,q,s)} \right)^2}{4 \, \mu_{\mathsmaller{\chi^2}}}
    \end{gather*}
    Finally, returning to Equation \ref{eq:constvar1} obtains the desired result:
    \begin{align*}
        \lim_{d\to\infty} \Var \left[ \norm{X}_2^T \right]
        &= n \left( \frac{\left( \sigma_{\mathsmaller{\chi^2}}\right)^2}{4 \, \mu_{\mathsmaller{\chi^2}}}  + 2 \lim_{d\to\infty} \Cov \left( \norm{X}_2^{(p,q)}, \norm{X}_2^{(q,s)}  \right) \right) \\ &= n \left( \frac{\left( \sigma_{\mathsmaller{\chi^2}}\right)^2}{4 \, \mu_{\mathsmaller{\chi^2}}}  + 2 \frac{\left( \varphi^{(p,q,s)} \right)^2}{4 \, \mu_{\mathsmaller{\chi^2}}} \right) \\ &= n \left( \frac{\left( \sigma_{\mathsmaller{\chi^2}}\right)^2 + 2 \left( \varphi^{(p,q,s)} \right)^2}{4 \, \mu_{\mathsmaller{\chi^2}}} \right)
    \end{align*}
\end{proof}

We can then evaluate \(\left( \sigma_{\mathsmaller{\chi^2}}\right)^2\) and \(\left( \varphi^{(p,q,s)} \right)^2\) for the uniform coordinate sampling used in this paper, like we did for \(\mu_{\mathsmaller{\chi^2}}\) after proving Lemma \ref{proof:boundd}:
\begin{align*}
    \left( \sigma_{\mathsmaller{\chi^2}}\right)^2 =& \int_0^1 \int_0^1 f(\omega) f(\rho) ((\rho - \omega)^2 - \mu_{\mathsmaller{\chi^2}})^2 d\rho \, d\omega = \int_0^1 \int_0^1 \left((\rho - \omega)^2 - \frac{1}{6}\right)^2 d\rho \, d\omega = \frac{7}{180} \\ \left( \varphi^{(p,q,s)} \right)^2 &= \int_0^1 \int_0^1 \int_0^1  f(\tau) f(\omega) f(\rho) ((\rho - \omega)^2 - \mu_{\mathsmaller{\chi^2}})((\omega - \tau)^2 - \mu_{\mathsmaller{\chi^2}}) d\rho \, d\omega \, d\tau \\ &= \int_0^1 \int_0^1 \int_0^1 \left((\rho - \omega)^2 - \frac{1}{6}\right)\left((\omega - \tau)^2 - \frac{1}{6}\right) d\rho \, d\omega \, d\tau \\ &= \frac{1}{180}
\end{align*}
Plugging in values yields:
\begin{gather*}
    \lim_{d\to\infty} \Var \left[ \norm{X}_2^T \right] = \frac{3}{40} n
\end{gather*}
Evaluating for 10 and 20 nodes, respectively, we obtain variance limits of \(\frac{3}{4}\) and \(\frac{3}{2}\), translating to standard deviation limits of approximately \(0.87\) and \(1.22\). Our empirical results for standard deviation of random tour length in Figure \ref{fig:rawstd} very closely align.

\phantomsection
\addcontentsline{toc}{subsection}{Theorem 6}
\vspace{1\baselineskip}
\begin{theorem}
    The difference between expected random tour length and expected optimal tour length (random tour suboptimality) approaches a constant value as \(d \to \infty\).
    \label{proof:constspan}
\end{theorem}
\begin{proof}
    Here we leverage an existing derivation that bounds the limit of expected absolute contrast for a fixed number of nodes. Taking the norm from an arbitrary query point to each node coordinate, absolute contrast is defined as the maximum difference between norms. Specifying the Euclidean case for Corollary 2 in Aggarwal et al. \cite{aggarwal2001surprising}, the bound can be formally described as follows:
    \begin{gather*}
        \lambda \leq \lim_{d\to\infty} \E \left[ \max_{u\in Q} \norm{X}_2^{(p,u)} - \min_{v\in Q} \norm{X}_2^{(p,v)} \right] \leq (n - 2) \lambda \numberthis \label{eq:contrast}
    \end{gather*}
    where node \(p\) is our chosen query point,\footnote{Aggarwal et al. consistently use the origin as their query point but without loss of generality. Our upper bound term is scaled by \((n-2)\) to account for removing node \(p\) from the end point set \(Q\).} \(Q\) is the set of remaining nodes such that \(p \notin Q\), and \(\lambda\) is a non-negative constant. This bound applies to any node coordinate distribution satisfying i.i.d. sampling for each dimension.

    Let \(\norm{X}_2^{T^*}\) be defined as the length of the optimal tour between the sampled nodes, that which incurs minimum Euclidean distance:
    \begin{gather*}
        \norm{X}_2^{T^*} = \sum_{k=1}^n \norm{X}_2^{(p_k,q_k^*)} \quad \text{s.t. } p_k = q_{k-1}^*,\; p_1 = q_n^*
    \end{gather*}
    where node \(q_k^*\) forms the optimal subsequent edge from node \(p_k\). Random tour suboptimality can then be expressed as \(\E \left[ \norm{X}_2^{T} - \norm{X}_2^{T^*}\right]\), the expected difference between random and optimal tour length.

    First, we extend Aggarwal et al.'s Corollary 2 to upper bound the limit of random tour suboptimality. Let \(q \in Q\) be a randomly selected node and let \(q^* \in Q\) form \((p,q^*)\), an optimal edge.\footnote{Each node \(p\) has two alternatives for \(q^*\), but this choice does not affect the analysis.} It follows that
    \begin{align*}
        \norm{X}_2^{(p,q)} - \norm{X}_2^{(p,q^*)} &\leq \max_{u\in Q} \norm{X}_2^{(p,u)} - \min_{v\in Q} \norm{X}_2^{(p,v)} \\ \lim_{d\to\infty} \E \left[ \norm{X}_2^{(p,q)} - \norm{X}_2^{(p,q^*)} \right] &\leq \lim_{d\to\infty} \E \left[ \max_{u\in Q} \norm{X}_2^{(p,u)} - \min_{v\in Q} \norm{X}_2^{(p,v)} \right] \\ &\leq (n - 2) \lambda \quad \text{via Rel. \ref{eq:contrast}}
    \end{align*}
    Repeating this reasoning with each node taking the perspective as the query point \(p\) then summing all inequalities yields:
    \begin{gather*}
        \sum_{k=1}^n \lim_{d\to\infty} \E \left[ \norm{X}_2^{(p_k,q_k)} - \norm{X}_2^{(p_k,q_k^*)} \right] \leq n (n-2) \lambda
    \end{gather*}
    Because we know each limit inside the sum does not diverge,\footnote{We can easily show each limit is lower bounded using the mirroring relation: \\ \(\norm{X}_2^{(p,q)} - \norm{X}_2^{(p,q^*)} \geq \min\limits_{v\in Q} \norm{X}_2^{(p,v)} - \max\limits_{u\in Q} \norm{X}_2^{(p,u)} \)} we can bring the limit outside the sum:
    \begin{align*}
        \lim_{d\to\infty} \sum_{k=1}^n \E \left[ \norm{X}_2^{(p_k,q_k)} - \norm{X}_2^{(p_k,q_k^*)} \right] &\leq n (n-2) \lambda \\ \lim_{d\to\infty} \left( \sum_{k=1}^n \E \left[ \norm{X}_2^{(p_k,q_k)} \right] - \sum_{j=1}^n \E \left[ \norm{X}_2^{(p_j,q_j^*)} \right] \right) &\leq n (n-2) \lambda
    \end{align*}
    Like each random edge, each optimal edge variable \(\norm{X}_2^{(p_j,q_j^*)}\) is identically distributed given the choice of starting node \(p_1\) is arbitrary. Therefore, the above relation is equivalent to
    \begin{align*}
        \lim_{d\to\infty} \left( \E \left[ \norm{X}_2^{T} \right] - \E \left[ \norm{X}_2^{T^*}\right] \right) &\leq n (n-2) \lambda \\ \lim_{d\to\infty} \E \left[ \norm{X}_2^{T} - \norm{X}_2^{T^*}\right] &\leq n (n-2) \lambda
    \end{align*}
    showing the desired upper bound. From here it is straightforward to show a non-negative lower bound. By the definition of optimality,
    \begin{align*}
        \norm{X}_2^{T^*} &\leq \norm{X}_2^{T} \\ 0 &\leq \norm{X}_2^{T} - \norm{X}_2^{T^*} \\ 0 &\leq \lim_{d\to\infty} \E \left[ \norm{X}_2^{T} - \norm{X}_2^{T^*} \right]
    \end{align*}
    Thus,
    \begin{gather*}
        0 \leq \lim_{d\to\infty} \E \left[ \norm{X}_2^{T} - \norm{X}_2^{T^*} \right] \leq n (n-2) \lambda
    \end{gather*}
\end{proof}

In words, in the limit as \(d\to\infty\), random tour suboptimality approaches a non-negative constant value. The upper bound is dependent on the number of nodes and, through \(\lambda\), dependent on the coordinate sampling distribution. 

If one can show \(\E\left[ \norm{X}_2^{T} - \norm{X}_2^{T^*} \right]>0\) for arbitrary spatial dimensions, we could obtain the stricter bound
\begin{gather*}
    \epsilon \leq \lim_{d\to\infty} \E \left[ \norm{X}_2^{T} - \norm{X}_2^{T^*} \right] \leq n (n-2) \lambda
\end{gather*}
for some \(0 < \epsilon \leq n (n-2) \lambda\) where \(\epsilon\) may be a function of \(n\). For uniform coordinate sampling, this seems almost certainly true. Using Markov’s inequality, it would be sufficient to show that for arbitrary spatial dimensions
\begin{gather*}
    \mathbb{P} \left[ \left(\norm{X}_2^{T} - \norm{X}_2^{T^*} \right) \geq \epsilon \right] > 0
\end{gather*}
While intuitive, we forgo attempting a proof.


\end{document}